\documentclass[10pt,twocolumn,letterpaper]{article}

\usepackage{iccv}
\usepackage{times}
\usepackage{epsfig}
\usepackage{graphicx}
\usepackage{amsmath}
\usepackage{amssymb}
\usepackage{wrapfig}

\usepackage[space]{grffile}
\usepackage{booktabs}
\usepackage{multirow}
\usepackage{xcolor,colortbl}
\usepackage{bm}
\usepackage{cuted}
\usepackage{ulem}

\usepackage[accsupp]{axessibility}

\usepackage[breaklinks=true,bookmarks=false]{hyperref}

\newcommand{\revise}[1]{\textcolor{black}{#1}}

\iccvfinalcopy %

\def\iccvPaperID{****} %

\ificcvfinal\fi

\begin{document}

\title{Ponder: Point Cloud Pre-training via Neural Rendering}

\author{
Di Huang$^{1,2}$~~~ 
Sida Peng$^3$~~~
Tong He$^{2,\dag}$~~~
Honghui Yang$^{2,3}$~~~ 
Xiaowei Zhou$^{3}$~~~
Wanli Ouyang$^2$~~~
\vspace{0.2cm} \\
The University of Sydney$^1$~~~
Shanghai AI Laboratory$^2$~~~
Zhejiang University$^3$
\vspace{0.2cm}
}

\maketitle
\ificcvfinal\thispagestyle{empty}\fi

\begin{abstract}
We propose a novel approach to self-supervised learning of point cloud representations by differentiable neural rendering. Motivated by the fact that informative point cloud features should be able to encode rich geometry and appearance cues and render realistic images, we train a point-cloud encoder within a devised point-based neural renderer by comparing the rendered images with real images on massive RGB-D data. 
The learned point-cloud encoder can be easily integrated into various downstream tasks, including not only high-level tasks like 3D detection and segmentation but also low-level tasks like 3D reconstruction and image synthesis. 
Extensive experiments on various tasks demonstrate the superiority of our approach compared to existing pre-training methods. 

{\let\thefootnote\relax\footnotetext{$^{\dag}$denote corresponding author.}}
\end{abstract}
\section{Introduction}
\label{sec:introduction}

We have witnessed the widespread success of supervised learning in developing vision tasks, such as image classification~\cite{he2016deep, dosovitskiy2020image} and object detection~\cite{ren2015faster, he2017mask, MIR-2022-05-171}.
In contrast to the 2D image domain, current 3D point cloud benchmarks only maintain limited annotations, in terms of quantity and diversity, due to the extremely high cost of laborious labeling. Self-supervised learning (SSL) for point cloud~\cite{xie2020pointcontrast, hou2021exploring, jiang2021guided, huang2021spatio, chen2022_4dcontrast, rao2021randomrooms, zhang_depth_contrast, wang2021unsupervised, yu2021pointbert, yan2022implicit, pang2022masked, liu2022masked, zhang2022point, min2022voxel}, consequently, becomes one of the main driving forces and has attracted increasing attention in the 3D research community. 

\begin{figure}[tb]
  \centering
  \includegraphics[width=1.0\linewidth]{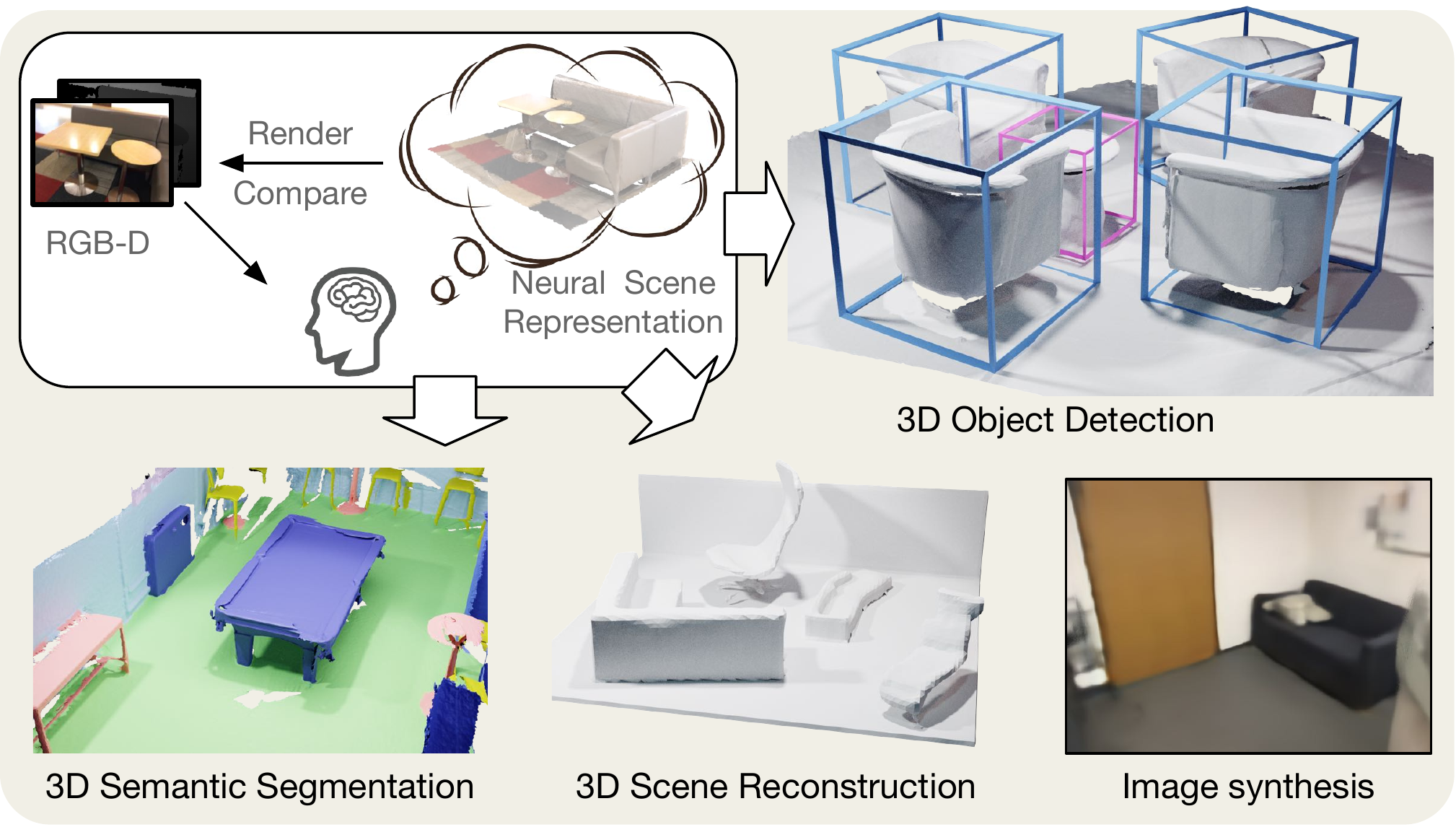}
  \caption{This work proposes a novel point cloud pre-training method via neural rendering, named \textbf{Ponder}. \textbf{Ponder} is directly trained with RGB-D image supervision, and can be used for various applications, e.g. 3D object detection, 3D semantic segmentation, 3d scene reconstruction, and image synthesis.}
  \label{fig:teaser tasks}
\end{figure}

Previous SSL methods for learning effective 3D representation can be roughly categorized into two groups: contrast-based~\cite{xie2020pointcontrast, hou2021exploring, jiang2021guided, huang2021spatio, chen2022_4dcontrast, rao2021randomrooms, zhang_depth_contrast} and completion-based~\cite{wang2021unsupervised, yu2021pointbert, yan2022implicit, pang2022masked, liu2022masked, zhang2022point, min2022voxel}.
Contrast-based methods are designed to maintain invariant representation under different transformations. 
To achieve this, informative samples are required. In the 2D image domain, the above challenge is addressed by (1) introducing efficient positive/negative sampling methods, (2) using a large batch size and storing representative samples, and (3) applying various data augmentation policies.
Inspired by these works, many works~\cite{xie2020pointcontrast, hou2021exploring, jiang2021guided, huang2021spatio, chen2022_4dcontrast, rao2021randomrooms, zhang_depth_contrast} are proposed to learn geometry-invariant features on 3D point cloud. 

Completion-based methods are another line of research for 3D SSL, which utilizes a pre-training task of reconstructing the masked point cloud based on partial observations. 
By maintaining a high masking ratio, 
such a simple task encourages the model to learn a holistic understanding of the input beyond low-level statistics. 
Although the masked autoencoders have been successfully applied for SSL in images~\cite{he2022masked} and videos~\cite{feichtenhofer2022masked, tong2022videomae}, it remains challenging and still in exploration due to the inherent irregularity and sparsity of the point cloud data. 

Different from the two groups of methods above, we propose \textbf{po}int cloud pre-training via neural re\textbf{nder}ing (Ponder).
Our motivation is that neural rendering, one of the most amazing progress and domain-specific design in 3D vision, can be leveraged to enforce the point cloud features being able to encode rich geometry and appearance cues.
As illustrated in Figure~\ref{fig:teaser tasks}, we
address the task of learning representative 3D features via point cloud rendering.
To the best of our knowledge, this is the first exploration of neural rendering for pre-training 3D point cloud models.
Specifically, given one or a sequence of RGB-D images, we lift them to 3D space and obtain a set of colored points.
Points are then forwarded to a 3D encoder to learn the geometry and appearance of the scene via a neural representation.
Provided specific parameters of the camera and the neural representation from the encoder, neural rendering is leveraged to render the RGB and depth images in a differentiable way. The network is trained to minimize the difference between rendered and observed 2D images. 
In doing so, our approach enjoys multiple advantages:
\begin{itemize}
    \item Our method is able to learn effective point cloud representation, which encodes rich geometry and appearance clues by leveraging neural rendering.

    \item Our method can be flexibly integrated into various tasks. 
    For the first time, we validate the effectiveness of the proposed pre-training method for low-level tasks like surface reconstruction and image synthesis tasks.
    
    \item The proposed method can leverage rich RGB-D images for pre-training. The easier accessibility of the RGB-D data enables the possibility of 3D pre-training on a large amount of data.
\end{itemize}
Our approach \revise{proposes a novel pretext task that} can serve as a strong alternative to contrast-based methods and completion-based methods in 3D point cloud pre-training.
The proposed framework, \textbf{Ponder}, is capable of accommodating a variety of point cloud backbones, both point-based and voxel-based, and has been rigorously evaluated on a range of challenging 3D tasks, including object detection, semantic segmentation, reconstruction, and image synthesis.
The consistent improvements demonstrate the effectiveness of our proposed \textbf{Ponder}.

\begin{figure}[tb]
  \centering
  \includegraphics[width=1.0\linewidth]{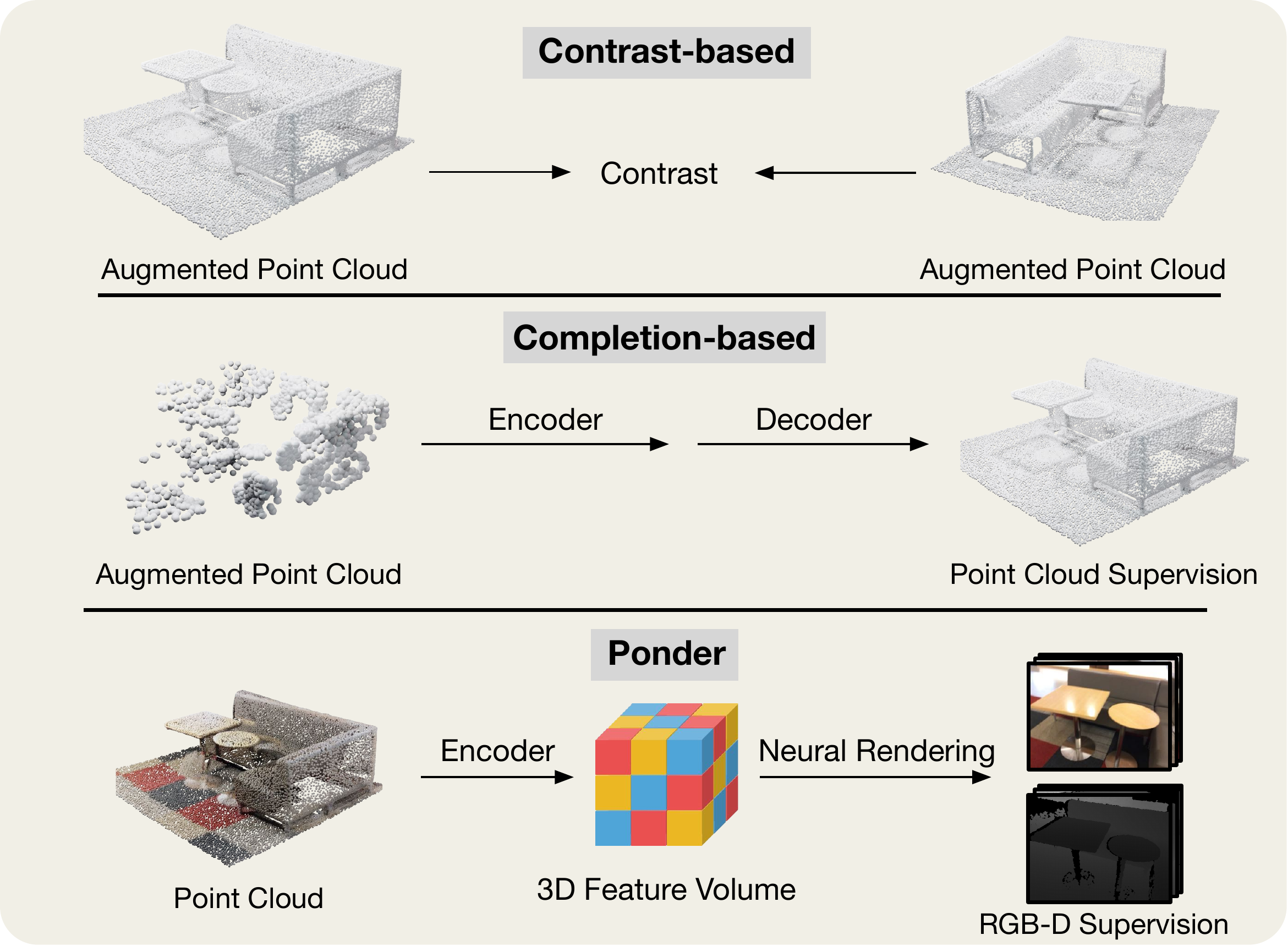}
  \caption{Different types of point cloud pre-training.
  }
   \label{fig:teaser methods}
\end{figure}

\section{Related Work}
\label{sec:related work}

\noindent \textbf{Neural rendering.} 
Neural Rendering is a type of rendering technology that uses neural networks to differentiablely render images from 3D scene representation.
NeRF\cite{mildenhall2021nerf} is one of the representative neural rendering methods, which represents the scene as the neural radiance field and renders the images via volume rendering.
Based on NeRF, there are a series of works~\cite{mueller2022instant, yu2021plenoctrees, wang2021neus, oechsle2021unisurf, Yu20arxiv_pixelNeRF, wang2021ibrnet, Zhang20arxiv_nerf++, reiser2021kilonerf, barron2021mip, MIR-2022-06-204} trying to improve the NeRF representation, including accelerate NeRF training, boost the quality of geometry, and so on.
Another type of neural rendering leverages neural point clouds as the scene representation. 
\cite{aliev2020neural, rakhimov2022npbg++} take points locations and corresponding descriptors as input, rasterize the points with z-buffer, and use a rendering network to get the final image. 
Later work of PointNeRF\cite{xu2022point} renders realistic images from neural point cloud representation using a NeRF-like rendering process. 
Our work is inspired by the recent progress of neural rendering.

\vspace{6 pt}
\noindent \textbf{Self-supervised learning in point clouds.}
Current methods can be roughly categorized into two categories: contrast-based and completion-based. Inspired by the works~\cite{he2020momentum, chen2020simple} from the 2D image domain, PointContrast~\cite{xie2020pointcontrast} is one of the pioneering works for 3D contrastive learning. Similarly, it encourages the network to learn invariant 3D representation under different transformations. 
Some works~\cite{hou2021exploring, jiang2021guided, huang2021spatio, chen2022_4dcontrast, rao2021randomrooms, zhang_depth_contrast} follow the pipeline by either devising new sampling strategies to select informative positive/negative training pairs, or explore various types of data augmentations.
Another line of work is completion-based~\cite{yu2021pointbert, yan2022implicit, pang2022masked, liu2022masked, zhang2022point, min2022voxel} methods, 
which get inspiration from Masked Autoencoders~\cite{he2022masked}. 
PointMAE~\cite{pang2022masked} proposes restoring the masked points via a set-to-set Chamfer Distance. 
VoxelMAE~\cite{min2022voxel} instead recovers the underlying geometry by distinguishing if the voxel contains points. 
Another work MaskPoint\cite{liu2022masked} pre-train point cloud encoder by performing binary classification to check if a sampled point is occupied. 
Later, IAE~\cite{yan2022implicit} proposes to pre-train point cloud encoder by recovering continuous 3D geometry in an implicit manner. 
Different from the above pipelines, we propose a novel framework for point cloud pre-training via neural rendering.

\vspace{6 pt}
\noindent \textbf{Multi-modal point cloud pre-training.}
Some recent works explore the pre-training pipeline with multi-modality data of 2D images and 3D point clouds. Pri3D\cite{hou2021pri3d} use 3D point cloud and multi-view images to pre-train the 2D image networks. CrossPoint\cite{afham2022crosspoint} aligns the 2D image features and 3D point cloud features through a contrastive learning pipeline. 
\cite{li2022closer} proposes a unified framework for exploring the invariances with different input data formats, including 2D images and 3D point clouds.
Different from previous methods, most of which attempt to align 2D images and 3D point clouds in the feature space, our method proposes to connect 2D and 3D in the RGB-D image domain via differentiable rendering.

\begin{figure*}[htb]
  \centering
  \includegraphics[width=1.0\linewidth]{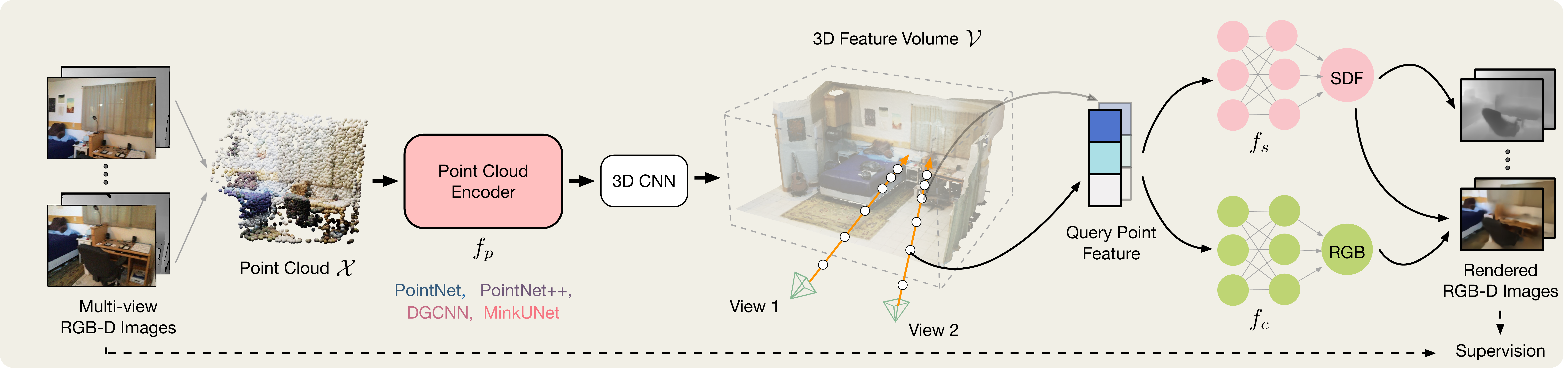}
  \caption{\textbf{The pipeline of our point cloud pre-training via neural rendering (Ponder).} 
  Given multi-view RGB-D images, we first construct the point cloud by back-projection, 
  then use a point cloud encoder $f_p$ to extract per-point features $\mathcal{E}$. 
  \revise{$\mathcal{E}$ are organized to a 3D feature volume by average pooling and then processed by the 3D convolution layer.}
  Finally, the 3D feature volume is rendered to multi-view RGB-D images via a differentiable neural rendering, which are compared with the input multi-view RGB-D images as the supervision. 
  }
   \label{fig:pipeline}
   \vspace{-0.3cm}
\end{figure*}

\section{Methods}
\label{sec:methods}
An overview of our \textbf{Ponder} is presented in Figure \ref{fig:pipeline}. 
Provided the camera pose, 3D point clouds are obtained by projecting the RGB-D images back to 3D space (Section \ref{sec: methods pre-processing}). Then, we extract point-wise feature using a point cloud encoder (Section \ref{sec: methods encoder}) and organize it to a 3D feature volume (Section \ref{sec: feature volume}), which is used to reconstruct the neural scene representation and render images in a differentiable manner (Section \ref{sec: methods radiance field reconstruction}).
\subsection{Constructing point cloud from RGB-D images}
\label{sec: methods pre-processing}

The proposed method makes use of sequential RGB-D images $\{(I_i, D_i)\}_{i=1}^N$, the camera intrinsic parameters $\{\textbf{K}_i\}_{i=1}^N$, and extrinsic poses $\{\bm{\xi}_i\}_{i=1}^N \in \textbf{SE}(3)$. $N$ is the input view number. $\textbf{SE}(3)$ refers to the Special Euclidean Group representing 3D rotations
and translations.
The camera parameters can be easily obtained from SfM or SLAM.

We construct the point cloud $\mathcal{X}$ by back-projecting  RGB-D images to point clouds in a unified coordinate:

\begin{equation}
    \mathcal{X} = \bigcup_i^N \pi^{-1} (I_i, D_i, \bm{\xi}_i, \textbf{K}_i) \text{,}
\end{equation}
where $\pi^{-1}$ back-projects the RGB-D image to 3D world space using camera poses. 
Note that different from previous methods which only consider the point location, our method attributes each point with both point location and RGB color. 
The details of $\pi^{-1}$ are provided in the supplementary material.

\subsection{Point cloud encoder for feature extraction} 
\label{sec: methods encoder}
Given the point cloud $\mathcal{X}$ constructed from RGB-D images, a point cloud encoder $f_p$ is used to extract per-point feature embedding $\mathcal{E}$:
\begin{equation}
    \mathcal{E} = f_p(\mathcal{X}) \text{.}
\end{equation}
The encoder $f_p$ pre-trained with the method mentioned in the Section~\ref{sec: methods radiance field reconstruction} serves as a good initialization for various downstream tasks.

\subsection{Building feature volume} 
\label{sec: feature volume}
\revise{
After completing feature extraction, we use average pooling to convert the point embeddings $\mathcal{E}$ into a 3D feature volume. We then employ a U-Net style 3D CNN to fill in the empty space and aggregate features from the surrounding points to obtain a dense 3D volume, denoted as $\mathcal{V}$. 
}

\subsection{Pre-training with Neural Rendering}
\label{sec: methods radiance field reconstruction}

This section introduces how to reconstruct the implicit scene representation and render images differentiablely.
We first give a brief introduction to neural scene representation, then illustrate how to integrate it into our point cloud pre-training pipeline. 
Last, we show the differentiable rendering formulation to render color and depth images from the neural scene representation.

\paragraph{Brief introduction of neural scene representation.}
Neural scene representation aims to represent the scene geometry and appearance through a neural network.
In this paper, we use the Signed Distance Function (SDF), which measures the distance between a query point and the surface boundary, to represent the scene geometry implicitly. 
SDF is capable of representing high-quality geometry details. 
For any query point of the scene, the neural network takes points features as input and outputs the corresponding SDF value and RGB value. In this way, the neural network captures both the geometry and appearance information of a specific scene.
Following NeuS\cite{wang2021neus}, the scene can be reconstructed as:
\begin{equation}
  s(\textbf{p}) = \tilde{f}_s(\textbf{p}), \quad c(\textbf{p}, \textbf{d}) = \tilde{f}_c(\textbf{p}, \textbf{d}),
  \label{eq:brief neural scene representation}
\end{equation}
where $\tilde{f}_s$ is the SDF decoder and $\tilde{f}_c$ is the RGB color decoder. 
$\tilde{f}_s$ takes point location $\textbf{p}$ as input, and predicts the SDF value $s$.
$\tilde{f}_c$ takes point location $\textbf{p}$ and viewing direction $\textbf{d}$ as input, and outputs the RGB color value $c$.
Both $\tilde{f}_s$ and $\tilde{f}_c$ are implemented by simple MLP networks. 

\paragraph{Neural scene representation from point cloud input in Ponder.}
To predict a neural scene representation from the input point cloud, we change the scene formulation to take 3D feature volume $\mathcal{V}$ as an additional input.
Specifically, given a 3D query point $\textbf{p}$ and viewing direction $\textbf{d}$, the feature embedding $\mathcal{V}(\textbf{p})$ can be extracted from the processed feature volume $\mathcal{V}$ by trilinear interpolation. 
The scene is then represented as:

\begin{equation}
  s(\textbf{p}) = f_s(\textbf{p}, \mathcal{V}(\textbf{p})), \quad c(\textbf{p}, \textbf{d}) = f_c(\textbf{p}, \textbf{d}, \mathcal{V}(\textbf{p})) \text{,}
  \label{eq:ponder neural scene representation}
\end{equation}
where $\mathcal{V}$ is predicted by the point cloud encoder $f_p$ and encodes information of each scene.
$f_s$ and $f_c$ are SDF and RGB decoders shared for all scenes.
Different from Equation (\ref{eq:brief neural scene representation}), which is used for storing single-scene information in the $\{\tilde{f}_s, \tilde{f}_c\}$, the formulation in Equation (\ref{eq:ponder neural scene representation}) includes an extra input $\mathcal{V}(\textbf{p})$
to facilitate representing the information of multiple scenes.

\paragraph{Differentiable rendering.}
Given the dense 3D volume $\mathcal{V}$ and viewing point, we use differentiable volume rendering to render the projected color images and depth images. 
For each rendering ray with camera origin $\textbf{o}$ and viewing direction $\textbf{d}$, we sample a set of ray points $\{\textbf{p}(z)|\textbf{p}(z) = \textbf{o} + z\textbf{d},z \in [z_n,z_f]\}$ along the ray, where $z$ denotes the length of the ray. Note that $\textbf{o}$ and $\textbf{d}$ can be calculated from paired camera parameters $\{(\bm{K}_i,  \bm{\xi}_i)\}$. $z_n$ and $z_f$ denote the near and far bounds of the ray. 
Different from previous methods~\cite{mildenhall2021nerf, wang2021neus}, we automatically determine $\{z_n, z_f\}$ by the ray intersection with the 3D feature volume box, using axis-aligned bounding boxes (AABB) algorithm.
Then, the ray color and depth value can be aggregated as:
\begin{align}
    \hat{C} &= \int_{z_n}^{z_f} \omega (z) c(\textbf{p}(z), \textbf{d})dz \text{,} \label{eq:render} \\
    \hat{D} &= \int_{z_n}^{z_f} \omega (z) z dz \text{,}
    \label{eq:render2}
\end{align}
where the $\hat{C}$ is the ray color and the $\hat{D}$ is the ray depth.
We follow NeuS\cite{wang2021neus} to build an unbiased and occlusion-awareness weight function $w(z)$: 
\begin{align}
    w(z) &= T(z) \cdot \rho(z) \text{.}
    \label{eq:neus}
\end{align}
$T(z)$ measures the accumulated transmittance from $z_n$ to $z$ and $\rho(z)$ is the occupied density function which are defined as:
\begin{align}
    T(z) &= \exp(-\int_{z_n}^{z_f}\rho(z)dz) \text{,}  \\
    \rho(z) &= \max \bigg (\frac{-\frac{d\Phi_h}{dz}(s(\textbf{p}(z)))}{\Phi_h(s(\textbf{p}(z)))}, 0 \bigg) \text{.}
    \label{eq:neus}
\end{align}
$\Phi_h (x)$ is the Sigmoid function $\Phi_h (x) = (1+e^{-hx})^{-1}$
where $h^{-1}$ is treated as a trainable parameter, $h^{-1}$ approaches to zero as the network training converges. 
In practice, we use a numerically approximated version by quadrature. 
We make the decode networks $\{f_s, f_c\}$ relatively smaller than ~\cite{mildenhall2021nerf, wang2021neus} to accelerate the training process.

\paragraph{Rendered examples.}
\begin{figure}[tb]
  \centering
  \includegraphics[width=1.0\linewidth]{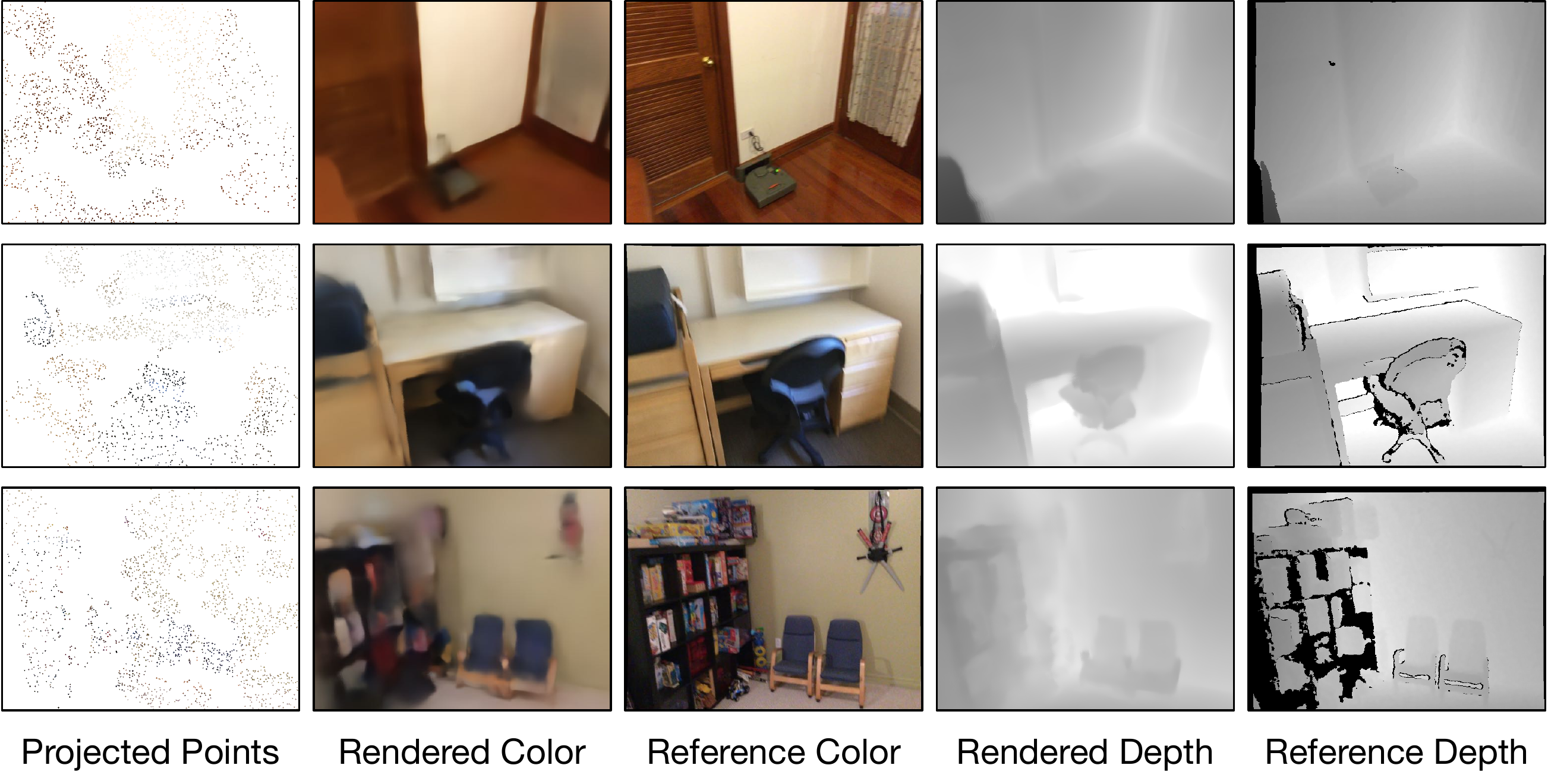}
  \caption{\textbf{Rendered images by Ponder} on the ScanNet validation set.
  The projected point clouds are visualized in the first column. 
  Even though input point clouds are very sparse, our model is still capable of rendering color and depth images similar to the reference images. }
   \label{fig:qualitative figure}
   \vspace{-0.3cm}
\end{figure}
The rendered color images and depth images are shown in Figure \ref{fig:qualitative figure}. 
As shown in the figure, even though the input point cloud is pretty sparse, our method is still capable of rendering color and depth images similar to the reference image.

\subsection{Pre-training loss} 
\label{sec: methods loss}
We leverage the input $\{I_i, D_i\}$ to supervise neural scene representation reconstruction.
The total loss function contains five parts,
\begin{equation}
  L = \lambda_c L_c + \lambda_d L_d + \lambda_e L_e + \lambda_s L_s + \lambda_f L_f ,
  \label{eq:loss}
\end{equation}
which are loss functions responsible for color supervision $L_c$, depth supervision $L_d$, Eikonal regularization $L_e$, near-surface SDF supervision $L_s$, and free space SDF supervision $L_f$. These loss functions are illustrated in the following section.

\paragraph{Color and depth loss.}
$L_c$ and $L_d$ are the color loss and depth loss, which measure consistency between the rendered pixels and the ground-truth pixels. Assume that we sample $N_r$ rays for each image and $N_p$ points for each ray, then the $L_c$ and $L_d$ can be written as:
\begin{align}
    L_c &= \frac{1}{N_r} \sum_i^{N_r} ||\hat{C} - C||_2^2  \\
    L_d &= \frac{1}{N_r} \sum_i^{N_r} ||\hat{D} - D||_2^2 \textbf{,}
    \label{eq:color depth}
\end{align}
where $C$ and $D$ are the ground-truth color and depth respectively for each ray,  $\hat{C}$ and $\hat{D}$ are their corresponding rendered ones in Eq.~(\ref{eq:render}) and Eq.~(\ref{eq:render2}). 

\paragraph{Loss for SDF regularization.}
$L_e$ is the widely used Eikonal loss~\cite{gropp2020implicit} for SDF regularization:
\begin{equation}
  L_{e} = \frac{1}{N_r N_p}\sum_{i,j}^{N_r, N_p}(|\nabla s(\textbf{p}_{i,j})|-1)^2 \text{,}
\end{equation}
where $\nabla s(\textbf{p}_{i,j})$ denotes the gradient of SDF $s$ at location $\textbf{p}_{i,j}$. Since SDF is a distance measure, $L_e$ encourages this distance to have a unit norm gradient at the query point.

\paragraph{Near-surface and free space loss for SDF.}
To stabilize the training and improve the reconstruction performance, similar to iSDF~\cite{Ortiz:etal:iSDF2022} and GO-Surf~\cite{wang2022go}, we add additional approximate SDF supervision to help the SDF estimation. Specifically, for near-surface points, the difference between rendered depth and ground-truth depth can be viewed as the pseudo-SDF ground-truth supervision; for points far from the surface, a free space loss is used to regularize the irregular SDF value additionally. 
To calculate the approximate SDF supervision, we first define an indicator $b(z)$ for each sampled ray point with ray length $z$ and corresponding GT depth $D$:
\begin{equation}
    b(z)=D-z \text{.}
\end{equation}
$b(z)$ can be viewed as the approximate SDF value, which is credible only when $b(z)$ is small.
Let $t$ be a human-defined threshold, which is set as 0.05 in this paper. For sampled ray points that satisfy $b(z) \leq t$, we leverage the near-surface SDF loss to constrain the SDF prediction $s(z_{i,j})$:
\begin{equation}
    L_s = \frac{1}{N_r N_p}\sum_{i,j}^{N_r, N_p} |s(z_{i,j}) - b(z_{i,j})| \text{.}
\end{equation}
For the remaining sampled ray points, we use a free space loss:
\begin{equation}
    L_f = \frac{1}{N_r N_p}\sum_{i,j}^{N_r, N_p} {\rm max}(0, e^{-\alpha \cdot s(z_{i,j})} - 1, s(z_{i,j}) - b(z_{i,j})) \text{,}
\end{equation}
where $\alpha$ is set as 5 following the same with \cite{Ortiz:etal:iSDF2022, wang2022go}.
Note that due to the noisy depth images, we only apply $L_s$ and $L_f$ on the rays that have valid depth values.

In our experiments, we follow a similar loss of weight with GO-Surf~\cite{wang2022go}, which sets $\lambda_c$ as 10.0, $\lambda_d$ as 1.0, $\lambda_s$ as 10.0, and $\lambda_f$ as 1.0. 
We observe that the Eikonal term in our method can easily lead to over-smooth reconstructions, thus we use a small weight of 0.01 for the Eikonal loss.
\section{Experiments}
\label{sec:experiments}

\subsection{Pre-training}

\noindent \textbf{Datasets.} 
We use ScanNet\cite{dai2017scannet} RGB-D images as our pre-training data. ScanNet is a widely used real-world indoor dataset, which contains more than 1500 indoor scenes. Each scene is carefully scanned by an RGB-D camera, leading to about 2.5 million RGB-D frames in total. We follow the same train/val split with VoteNet\cite{qi2019deep}.

\vspace{6 pt}
\noindent \textbf{Data preparation.} 
During pre-training, a mini-batch of batch size 8 includes point clouds from 8 scenes. The point cloud of a scene, serving as the input of the point cloud encoder in our approach, is back-projected from the 5 RGB-D frames of the video for the scene with an interval of 20. The 5 frames are also used as the supervision of the network.
We randomly down-sample the input point cloud to 20,000 points and follow the masking strategy as used in Mask Point~\cite{liu2022masked}. 

\vspace{6 pt}
\noindent \textbf{Implementation details.} 
We train the proposed pipeline for 100 epochs using an AdamW optimizer~\cite{loshchilov2017decoupled} with a weight decay of 0.05. 
The learning rate is initialized as 1e-4 with Exponential scheduling. 
For the rendering process, we randomly choose 128 rays for each image and sample 128 points for each ray.
More implementation details can be found in the supplementary materials.

\subsection{Transfer Learning}
In contrast to previous methods, our approach is able to encode rich geometry and appearance cues into the point cloud representations via neural rendering.
These strengths make it flexible to be applied to various tasks, including not only 3D semantic segmentation and 3D detection tasks but also low-level surface reconstruction and image synthesis.

\subsubsection{High-level 3D Tasks}

\paragraph{3D object detection.}

 \begin{table*}
  \small
  \centering
  \scalebox{1.05}{
  \begin{tabular}{l|c c c |c c c c}
    \toprule
    \multirow{2}{*}{Method} & Detection & Pre-training & Pre-training & \multicolumn{2}{c}{ScanNet} & \multicolumn{2}{c}{SUN RGB-D}\\
     & Model & Type & Epochs & $\rm {AP}_{50}$ $\uparrow$ & $\rm {AP}_{25}$ $\uparrow$ & $\rm {AP}_{50}$ $\uparrow$ & $\rm {AP}_{25}$ $\uparrow$\\
    \toprule
    3DETR\cite{misra2021-3detr} & 3DETR & - & - & 37.5 & 62.7 & 30.3 & 58.0 \\
    Point-BERT\cite{yu2021pointbert} & 3DETR & Completion & 300 & 38.3 & 61.0 & - & - \\
    MaskPoint\cite{liu2022masked} & 3DETR & Completion & 300 & 40.6 & 63.4 & - & - \\
    \midrule
    VoteNet \cite{qi2019deep} & VoteNet & - & - & 33.5 & 58.6 & 32.9 & 57.7 \\
    STRL\cite{huang2021spatio} & VoteNet & Contrast & 100 & 38.4 & 59.5 & 35.0 & 58.2 \\
    RandomRooms\cite{rao2021randomrooms} & VoteNet & Contrast & 300 & 36.2 & 61.3 & 35.4 & 59.2 \\
    PointContrast\cite{xie2020pointcontrast} & VoteNet & Contrast & - & 38.0 & 59.2 & 34.8 & 57.5 \\
    PC-FractalDB\cite{yamada2022point} & VoteNet & Contrast & - & 38.3 & 61.9 & 33.9 & 59.4 \\
    DepthContrast\cite{zhang_depth_contrast} & VoteNet & Contrast & 1000 & 39.1 & 62.1 & 35.4 & 60.4 \\
    IAE\cite{yan2022implicit} & VoteNet & Completion &  1000 & 39.8 & 61.5 & 36.0 & 60.4 \\
    \toprule
    \textbf{Ponder} & VoteNet & Rendering & 100 &  \textbf{41.0} (\textcolor{purple}{+7.5}) & 63.6 (\textcolor{purple}{+5.0}) & \textbf{36.6} (\textcolor{purple}{+3.7}) & \textbf{61.0} (\textcolor{purple}{+3.3}) \\
    \bottomrule
  \end{tabular}
  }
  \vspace{6pt}
  \caption{
  \textbf{3D object detection} $\textit{AP}_{25}$ \textit{and} $\textit{AP}_{50}$ {on ScanNet and SUN RGB-D.} 
  VoteNet\cite{qi2019deep} is a baseline model.
  \textcolor{purple}{Purple numbers} \revise{indicate improvements over the corresponding baseline.}
  The DepthContrast\cite{zhang_depth_contrast} and Point-BERT\cite{yu2021pointbert} results are adopted from IAE\cite{yan2022implicit} and MaskPoint\cite{liu2022masked}.
  \textbf{Ponder} outperforms both VoteNet-based and 3DETR-based point cloud pre-training methods with fewer training epochs.
  }
  \label{tab:3d object detection}
\end{table*}

\begin{table}[]
  \small
  \centering
  \begin{tabular}[t]{l | c c}
            \toprule
            Method & $AP_{50}$ $\uparrow$ & $AP_{25}$ $\uparrow$\\
            \midrule
            VoteNet\cite{qi2019deep} & 33.5 & 58.7 \\
            3DETR\cite{misra2021-3detr} & 37.5 & 62.7\\
            3DETR-m\cite{misra2021-3detr} & 47.0  & 65.0 \\
            H3DNet\cite{zhang2020h3dnet} & 48.1 & 67.2 \\
            \midrule
            \textbf{Ponder}+H3DNet & 50.9 (\textcolor{purple}{+2.8}) & 68.4 (\textcolor{purple}{+1.2}) \\
            \bottomrule
        \end{tabular}
  \vspace{6pt}
  \caption{
  \revise{
  \textbf{3D object detection}. $\textit{AP}_{25}$ \textit{and} $\textit{AP}_{50}$ {on ScanNet}. \textbf{Ponder} significantly boosts the detection accuracy of H3DNet by a margin of +2.8 and +1.2 for $\textit{AP}_{50}$ and $\textit{AP}_{25}$, respectively.
  }
  }
  \label{tab:3d detection h3dnet}
  \vspace{-0.3cm}
\end{table}
We select two representative approaches, Votenet~\cite{qi2019deep} and H3DNet~\cite{zhang2020h3dnet}, as the baselines. 
VoteNet leverages a voting mechanism to obtain object centers, which are used for generating 3D bounding box proposals.
\revise{By introducing a set of geometric primitives, H3DNet achieves a significant improvement in accuracy compared to previous methods.}
Two datasets are applied to verify the effectiveness of our method: ScanNet\cite{dai2017scannet} and SUN RGB-D\cite{song2015sun}.
Different from ScanNet, which contains fully reconstructed 3D scenes, 
SUN RGB-D is a single-view RGB-D dataset with 3D bounding box annotations. It has 10,335 RGB-D images for 37 object categories. 
For pre-training, we use PointNet++ as the point cloud encoder $f_p$, which is identical to the backbone used in VoteNet \revise{and H3DNet}.
We pre-train the point cloud encoder on the ScanNet dataset and transfer the weight as the downstream initialization.  
Following~\cite{qi2019deep}, we use average precision with 3D detection IoU threshold 0.25 and threshold 0.5 as the evaluation metrics.

The 3D detection results are shown in Table \ref{tab:3d object detection}. Our method improves the baseline of VoteNet without pre-training by a large margin, boosting $\rm AP_{50}$ by 7.5\% and 3.7\% for ScanNet and SUN RGB-D, respectively.
IAE~\cite{yan2022implicit} is a pre-training method that represents the inherent 3D geometry in a continuous manner. 
Our learned point cloud representation achieves higher accuracy because it is able to recover both the geometry and appearance of the scene. 
The $\rm AP_{50}$ and $\rm AP_{25}$ of our method are higher than that of IAE by 1.2\% and 2.1\% on ScanNet, respectively. 
\revise{
Besides, we have observed that our method surpasses the recent point cloud pre-training approach, MaskPoint~\cite{liu2022masked}, even when using a less sophisticated backbone (PointNet++ vs. 3DETR), as presented in Table \ref{tab:3d object detection}. 
To verify the effectiveness of \textbf{Ponder}, we also apply it for a much stronger baseline, H3DNet. As shown in Table \ref{tab:3d detection h3dnet}, our method surpasses H3DNet by +2.8 and +1.2 for $\rm AP_{50}$ and $\rm AP_{25}$, respectively.
}

\paragraph{3D semantic segmentation.}
3D semantic segmentation is another fundamental scene understanding task. 
\revise{
We select one of the top-performing backbones, MinkUNet\cite{choy20194d}, for transfer learning.
MinkUNet leverage 3D sparse convolution to extract effective 3D scene features.}
For pre-training, we use \revise{MinkUNet} as the point cloud encoder $f_p$, and pre-train the model on ScanNet.
\revise{
We report the finetuning results on the ScanNet dataset with the mean IoU of the validation set as the evaluation metric.
}
\revise{
Table \ref{tab:3D semantic segmentation mink} shows the quantitative results of \textbf{Ponder} with MinkUNet. The results demonstrate that \textbf{Ponder} is effective in improving the semantic segmentation performance, achieving a significant improvement of 1.3 mIoU. 
}

\subsubsection{Low-level 3D Tasks}
Low-level 3D tasks like scene reconstruction and image synthesis are getting increasing attention due to their wide applications. However, most of them are trained from scratch. How to pre-train a model with a good initialization is desperately needed. We are the first pre-training work to demonstrate a strong transferring ability to such low-level 3D tasks.

\paragraph{3D scene reconstruction.}
3D scene reconstruction task aims to recover the scene geometry, e.g. mesh, from the point cloud input. 
We choose ConvONet\cite{peng2020convolutional} as the baseline model, 
whose architecture is widely adopted in~\cite{chibane2020implicit, liu2020neural, yu2021plenoctrees}.
Following the same setting as ConvONet, we conduct experiments on the Synthetic Indoor Scene Dataset (SISD)\cite{peng2020convolutional}, which is a synthetic dataset and contains 5000 scenes with multiple ShapeNet~\cite{chang2015shapenet} objects. 
\revise{
To make a fair comparison with IAE~\cite{yan2022implicit}, we use the same VoteNet-style PointNet++ as the encoder of ConvONet, which down-samples the original point cloud to 1024 points. 
}
Following~\cite{peng2020convolutional}, we use Volumetric IoU, Normal Consistency \revise{(NC)}, and F-Score~\cite{tatarchenko2019single} with the threshold value of 1\% as the evaluation metrics.

\revise{
The results are shown in Table \ref{tab:3d scene reconstruction}. Compared to the baseline ConvONet model with PointNet++, IAE is not able to boost the reconstruction results, while the proposed approach can improve the reconstruction quality (+2.4\% for IoU). 
The results show the effectiveness of \textbf{Ponder} for the 3D reconstruction task.
}

\begin{table}[t]
\small
    \centering
    \begin{tabular}[t]{l|c}
    \toprule
    Method & mIoU $\uparrow$ \\
     \midrule
     PointNet++\cite{qi2017pointnet++} & 53.5 \\
     KPConv\cite{thomas2019KPConv} & 69.2 \\
     SparseConvNet\cite{3DSemanticSegmentationWithSubmanifoldSparseConvNet} & 69.3 \\
     PT\cite{zhao2021point} & 70.6  \\
     MinkUNet\cite{choy20194d} & 72.2  \\
     \midrule
     \textbf{Ponder}+MinkUNet &  \textbf{73.5} (\textcolor{purple}{+1.3})  \\
     \bottomrule
  \end{tabular}
  \vspace{6pt}
  \caption{
  \revise{
  \textbf{3D segmentation} \textit{mIoU} on ScanNet dataset. 
  }
  }
  \label{tab:3D semantic segmentation mink}
\end{table}

\begin{table}[t]
    \centering
    
    \scalebox{0.9}{
\begin{tabular}[t]{l| c c c c}
    \toprule
    Method & Encoder & IoU$\uparrow$ & NC$\uparrow$ & F-Score$\uparrow$ \\
    \toprule
    ConvONet\cite{peng2020convolutional} & PointNet++ & 77.8  & 88.7 & 90.6 \\
    IAE\cite{yan2022implicit} & PointNet++  & 75.7  & 88.7 & 91.0 \\
    \textbf{Ponder} & PointNet++ & \textbf{80.2} (\textcolor{purple}{+2.4}) & \textbf{89.3} & \textbf{92.0} \\
    \bottomrule
  \end{tabular}}
  \vspace{3pt}
  \caption{
  \textbf{3D scene reconstruction} \textit{IoU, NC, and F-Score} on SISD dataset \texttt{with PointNet++ model}. 
  \textbf{Ponder} is able to boost the reconstruction performance.
  }
  \label{tab:3d scene reconstruction}
  \vspace{-0.3cm}
\end{table}

\vspace{-0.3cm}
\paragraph{Image synthesis from point clouds.}

We also validate the effectiveness of our method on another low-level task of image synthesis from point clouds.  
We use Point-NeRF\cite{xu2022point} as the baseline. 
Point-NeRF uses neural 3D point clouds with associated neural features to render images. It can be used both for a generalizable setting for various scenes and a single-scene fitting setting. In our experiments, we mainly focus on the generalizable setting of Point-NeRF.
We replace the 2D image features of Point-NeRF with point features extracted by a DGCNN network. 
Following the same setting with PointNeRF, we use DTU\cite{jensen2014large} as the evaluation dataset. DTU dataset is a multiple-view stereo dataset containing 80 scenes with paired images and camera poses. 
We transfer both the DGCNN encoder and color decoder as the weight initialization of Point-NeRF. We use PSNR as the metric for synthesized image quality evaluation.

\begin{figure}[t]
  \centering
  \includegraphics[width=0.8\linewidth, , trim=0 40 0 60]{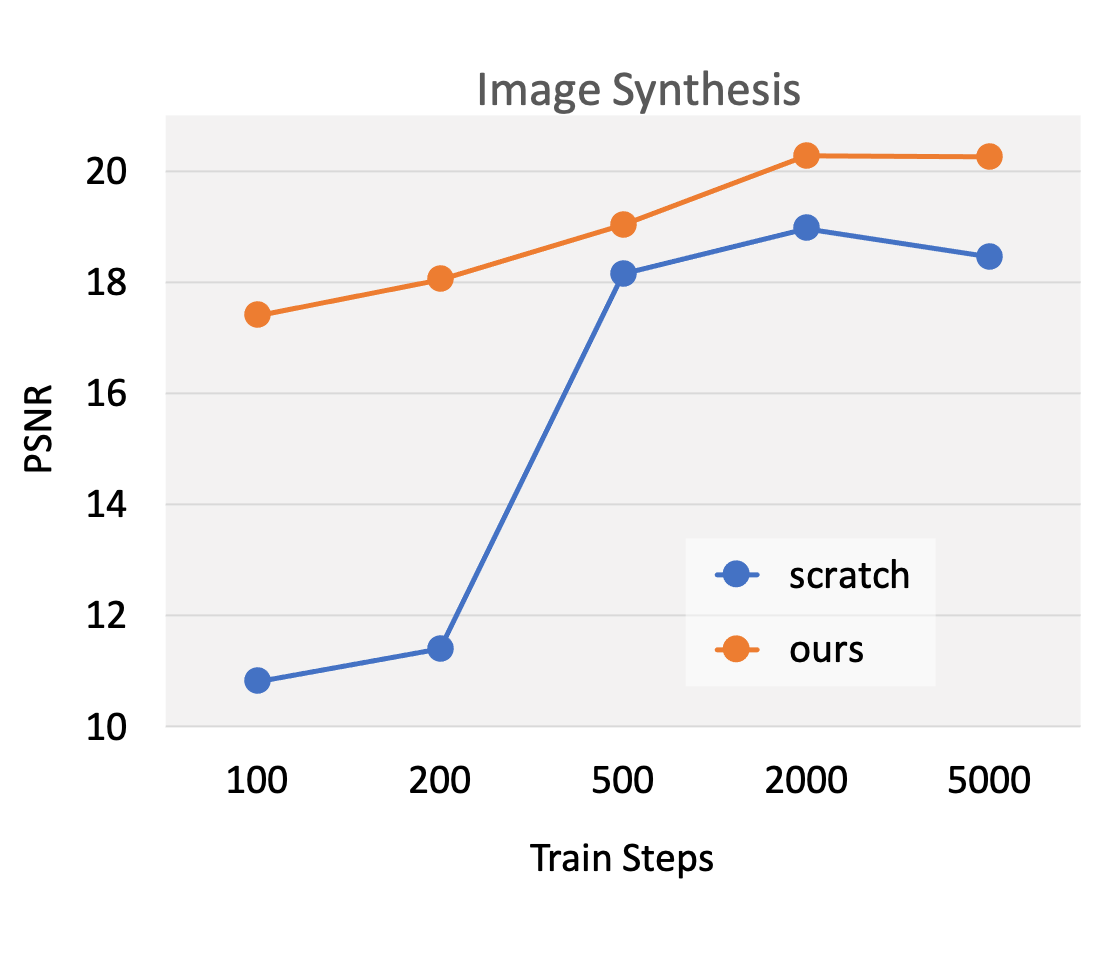}
  \caption{\textbf{\revise{Comparison of image synthesis from point clouds.}} Compared with training from scratch, 
  our \textbf{Ponder} model is able to converge faster and achieve better image synthesis results.
  }
  \label{fig:novel view}
  \vspace{-0.4cm}
\end{figure}

The results are shown in Figure \ref{fig:novel view}. 
By leveraging the pretrained weights of our method, the image synthesis model is able to converge faster with fewer training steps and achieve better final image quality than training from scratch.

\subsection{Ablation study}
\revise{
In this section, we conduct a series of ablation experiments to evaluate the effectiveness of our proposed approach. 
All experiments are conducted on ScanNet and SUN RGB-D datasets. We use 3D object detection for evaluation due to its simplicity.
}

\vspace{6 pt}
\noindent \textbf{Influence of Rendering Targets.} 
The rendering part of our method contains two items: RGB color image and depth image. We study the influence of each item with the transferring task of 3D detection. The results are presented in Table~\ref{tab:ablation study supervision}. 
Combining depth and color images for reconstruction shows the best detection results. In addition, using depth reconstruction presents better performance than color reconstruction for 3D detection.

\vspace{6 pt}
\noindent \textbf{Influence of mask ratio.} 
\revise{
To augment point cloud data, we employ random masking as one of the augmentation methods, which divides the input point cloud into 2048 groups with 64 points. 
In this ablation study, we evaluate the performance of our method with different mask ratios, ranging from 0\% to 90\%, on the ScanNet and SUN RGB-D datasets, and report the results in Table \ref{tab:supp mask ratio}. Notably, we find that even when no dividing and masking strategy is applied (0\%), our method achieves a competitive $\textit{AP}_{50}$ performance of 40.7 and 37.3 on ScanNet and SUN RGB-D, respectively.
Our method achieves the best performance on ScanNet with a mask ratio of 75\% and a $\textit{AP}_{50}$ performance of 41.7.
Overall, these results suggest that our method is robust to the hyper-parameter of mask ratio and can still achieve competitive performance without any mask operation.
}

\vspace{6 pt}
\noindent \textbf{Influence of 3D feature volume resolution.} 
\revise{
In our method, \textbf{Ponder} constructs a 3D feature volume with a resolution of [16, 32, 64], which is inspired by recent progress in multi-resolution 3D reconstruction. However, building such a high-resolution feature volume can consume significant GPU memory. To investigate the effect of feature volume resolution, we conduct experiments with different resolutions and report the results in Table \ref{tab:supp resolution}. From the results, we observe that even with a smaller feature volume resolution of 16, \textbf{Ponder} can still achieve competitive performance on downstream tasks.
}

\vspace{6 pt}

\begin{table}[htb]
\small
    \centering
  \begin{tabular}[t]{l| c c}
    \toprule
    Supervision & ScanNet   & SUN RGB-D \\
    \toprule
     VoteNet & 33.5 & 32.9 \\
     \midrule
     +Depth & 40.9 (\textcolor{purple}{+7.4})  & 36.1 (\textcolor{purple}{+3.2})  \\
     +Color & 40.5 (\textcolor{purple}{+7.0})  & 35.8 (\textcolor{purple}{+2.9}) \\
     \midrule
     +Depth+Color &  \textbf{41.0 (\textcolor{purple}{+7.5})}  & \textbf{36.6 (\textcolor{purple}{+3.7})}  \\
     \bottomrule
  \end{tabular}
  \vspace{6pt}
  \caption{
  \textbf{Ablation study for supervision type}.
  $\textit{3D detection}$ $\textit{{AP}}_{50}$ {on ScanNet and SUN RGB-D.} 
  Combining color supervision and depth supervision can lead to better detection performance than using a single type of supervision.
  }
  \label{tab:ablation study supervision}
\vspace{-0.3cm}
\end{table}

\begin{table}[htb]
    \small
    \centering
  \begin{tabular}[t]{c| c c}
    \toprule
    Mask ratio & ScanNet & SUN RGB-D\\
    \toprule
    VoteNet & 33.5 & 32.9 \\
    \midrule
    0\% & 40.7 (\textcolor{purple}{+7.2}) & \textbf{37.3 (\textcolor{purple}{+4.4})} \\
     \midrule
     25\% & 40.7 (\textcolor{purple}{+7.2})  & 36.2 (\textcolor{purple}{+3.3})  \\
     50\% & 40.3 (\textcolor{purple}{+6.8})  & 36.9 (\textcolor{purple}{+4.0})  \\
     75\% & \textbf{41.7 (\textcolor{purple}{+8.2})}  & 37.0 (\textcolor{purple}{+4.1})  \\
     90\% & 41.0 (\textcolor{purple}{+7.5})  & 36.6 (\textcolor{purple}{+3.7}) \\
     \bottomrule
  \end{tabular}
  \vspace{6pt}
  \caption{
  \textbf{Ablation study for mask ratio.} 
  $\textit{3D detection}$ $\textit{{AP}}_{50}$ {on ScanNet and SUN RGB-D.} 
  }
  \label{tab:supp mask ratio}
\vspace{-0.3cm}
\end{table}

\begin{table}[htb]
  \centering
  \small
  \begin{tabular}[t]{c| c c}
    \toprule
    Resolution & ScanNet & SUN RGB-D\\
     \toprule
     VoteNet & 33.5 & 32.9 \\
     \midrule
     16 & 40.7 (\textcolor{purple}{+7.2})  & \textbf{36.6 (\textcolor{purple}{+3.7})}  \\
     \midrule
     16+32+64 & \textbf{41.0 (\textcolor{purple}{+7.5})} & \textbf{36.6 (\textcolor{purple}{+3.7})} \\
     \bottomrule
  \end{tabular}
  \vspace{6pt}
  \caption{
  \textbf{Ablation study for feature volume resolution.} 
  $\textit{3D detection}$ $\textit{{AP}}_{50}$ {on ScanNet and SUN RGB-D.} 
  }
  \label{tab:supp resolution}
\vspace{-0.3cm}
\end{table}

\begin{table}[htb]
  \small
  \centering
  \begin{tabular}{c| c c}
    \toprule
    View number & ScanNet & SUN RGB-D\\
    \toprule
    VoteNet & 33.5 & 32.9 \\
     \midrule
     1 view & 40.1 (\textcolor{purple}{+6.6})  & 35.4 (\textcolor{purple}{+2.5})  \\
     3 views & 40.8 (\textcolor{purple}{+7.3})  & 36.0 (\textcolor{purple}{+3.1})  \\
     \midrule
     5 views &  \textbf{41.0 (\textcolor{purple}{+7.5})}  & \textbf{36.6 (\textcolor{purple}{+3.7})}  \\
     \bottomrule
  \end{tabular}
  \vspace{6pt}
  \caption{
  \textbf{Ablation study for view number.} 
  $\textit{3D detection}$ $\textit{{AP}}_{50}$ {on ScanNet and SUN RGB-D.} 
  Using multi-view supervision for point cloud pre-training can achieve better performance.
  }
  \label{tab:ablation study multiview}
\vspace{-0.3cm}
\end{table}

\noindent \textbf{Number of input RGB-D view.} 
Our method utilizes $N$ RGB-D images, where $N$ is the input view number. We study the influence of $N$ and conduct experiments on 3D detection, as shown in Table~\ref{tab:ablation study multiview}. We change the number of input views while keeping the scene number of a batch still 8.
Using multi-view supervision helps to reduce single-view ambiguity. Similar observations are also found in the multi-view reconstruction task~\cite{long2022sparseneus}. 
Compared with the single view, multiple views achieve higher accuracy, boosting $\rm AP_{50}$ by 0.9\% and 1.2\% for ScanNet and SUN RGB-D datasets, respectively. 

\begin{figure}[t]
  \centering
  \includegraphics[width=1.0\linewidth,trim=0 0 0 0]{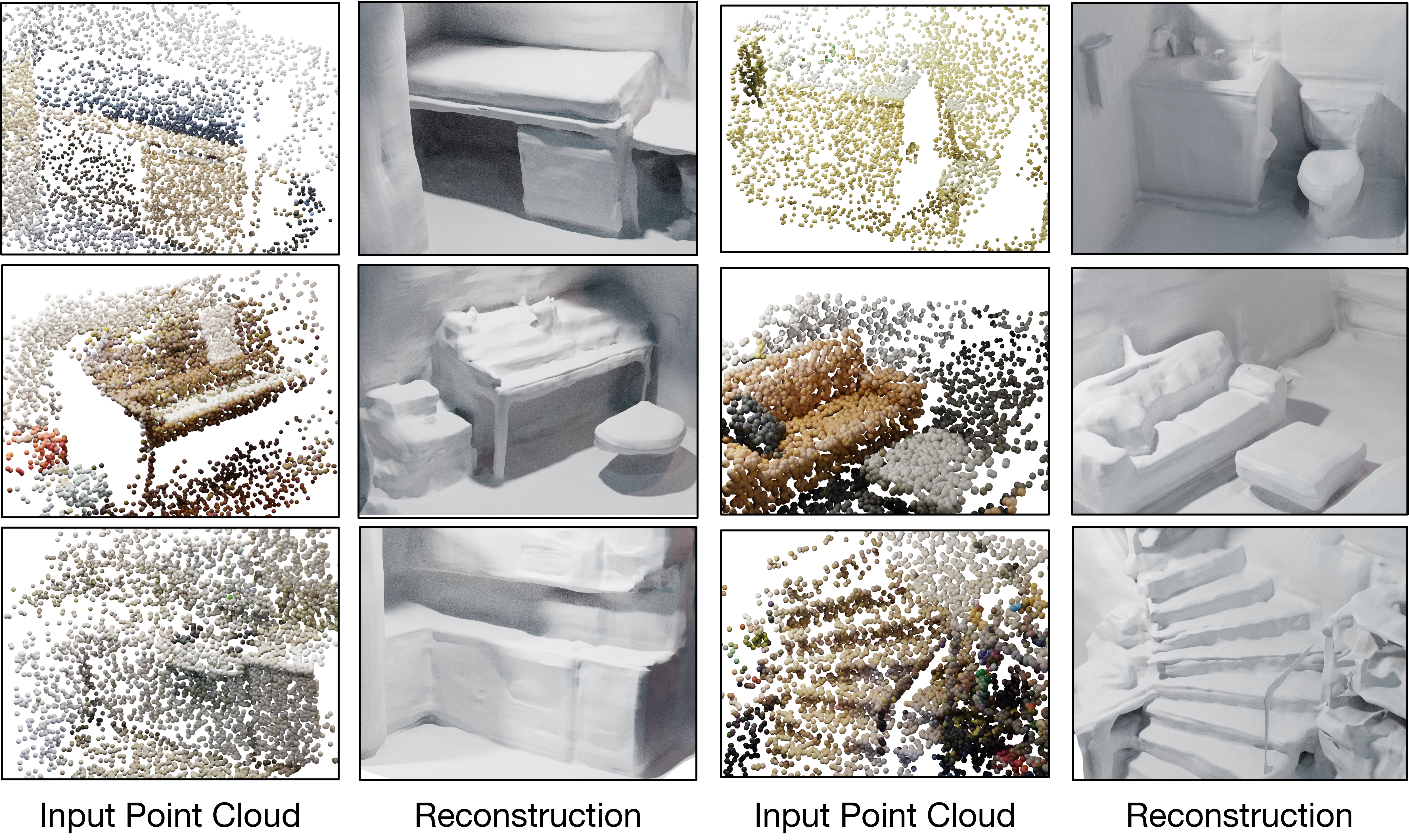}
  \caption{Reconstructed surface by \textbf{Ponder}. Our pre-training method can be easily integrated into the task of 3D reconstruction. 
  Despite the sparsity of the input point cloud (only 2\% points are used), our method can still recover precise geometric details.
}
  
   \label{fig:application reconstruction}
  \vspace{-0.3cm}
\end{figure}

\revise{
\subsection{Other applications}
The pre-trained model from our pipeline \textbf{Ponder} itself can also be directly used for surface reconstruction from sparse point clouds. 
Specifically, after learning the neural scene representation, we query the SDF value in the 3D space and leverage the Marching Cubes~\cite{lorensen1987marching} to extract the surface.
We show the reconstruction results in Figure \ref{fig:application reconstruction}. The results show that even though the input is sparse point clouds from complex scenes, our method is able to recover high-fidelity meshes.
Check the supplementary for more image synthesis and 3D reconstruction results.
}
\section{Conclusion}
\label{sec:conclusion}

This paper shows that differentiable neural rendering is a powerful tool for point cloud representation learning. 
The proposed pre-training pipeline, \textbf{Ponder}, is able to encode rich geometry and appearance cues into the point cloud representation via neural rendering. 
For the first time, our model can be transferred to both high-level 3D perception tasks and 3D low-level tasks, like 3D reconstruction and image synthesis from point clouds. 
Besides, the learned \textbf{Ponder} model can be directly used for 3D reconstruction and image synthesis from sparse point clouds. 
We also exploratively validate the effectiveness of \textbf{Ponder} on \textbf{outdoor scenario} and other input modalities, where we observe 1.41\% mAP improvement for 3D multiview object detection on the NuScene dataset that takes multiview images as input (details in the supplementary material). 

Several directions could be explored in future works.
\revise{First, recent progress in neural representations could help \textbf{Ponder} achieve better rendering quality and gain more accurate supervision from 2D images.
Second, thanks to the flexible architecture design, \textbf{Ponder} can potentially be expanded to other self-supervised learning fields, e.g., pre-training 2D image backbones, and other downstream tasks.
}

{\small
\bibliographystyle{ieee_fullname}
\bibliography{egbib}
}

\clearpage

\appendix

\begin{strip}
\begin{center}
\textbf{\Large Ponder: Point Cloud Pre-training via Neural Rendering}
\end{center}
\begin{center}\textbf{\Large Supplementary Material}
\end{center}
\vspace{20pt}
\end{strip}

\section{Implementation Details}
\label{sec: supp implementation detatils}
In this section, we give more implementation details of our \textbf{Ponder} model. 

\subsection{Pre-training Details}

\paragraph{Network architecture.} 
\revise{To process the extracted 3D feature volume, our approach utilizes a 3D U-Net. We adopt the standard implementation of 3D U-Net, which consists of four down-sampling stages with corresponding channels of 32, 64, 128, and 256, respectively.
All convolution layers use a 3D kernel of size 3. To construct the neural rendering decoders, Ponder employs a five-layer MLP network as the SDF decoder and a three-layer MLP network as the RGB decoder.}

\begin{figure}[h]
  \centering
   \includegraphics[width=0.9\linewidth]{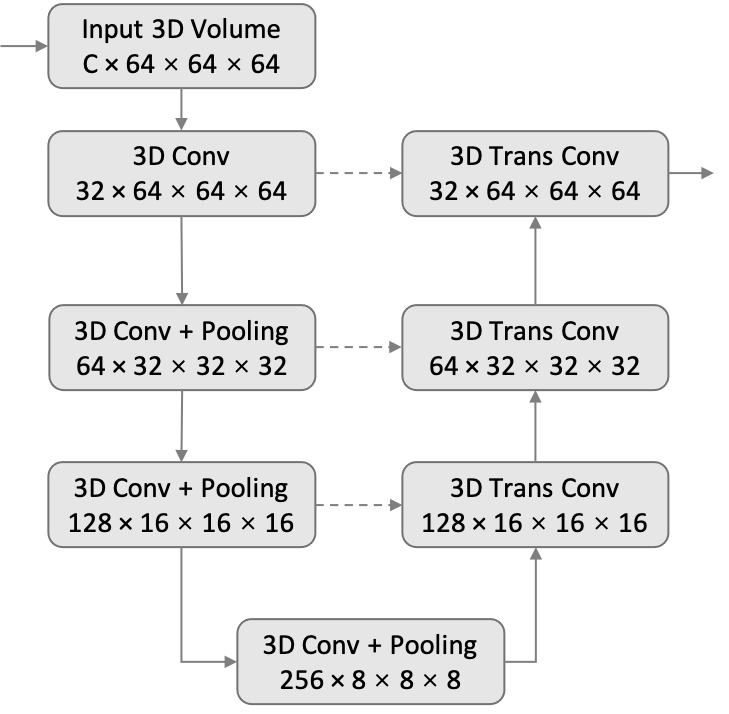}

   \caption{\textbf{3D U-Net architecture.}}
   \label{fig: network architecture}
   \vspace{-20pt}
\end{figure}

\paragraph{3D feature volume.} 
\setlength{\columnsep}{10pt}%
\setlength{\intextsep}{5pt}
\begin{wrapfigure}{r}{0.cm}
  \centering
  \includegraphics[width=0.17\textwidth]{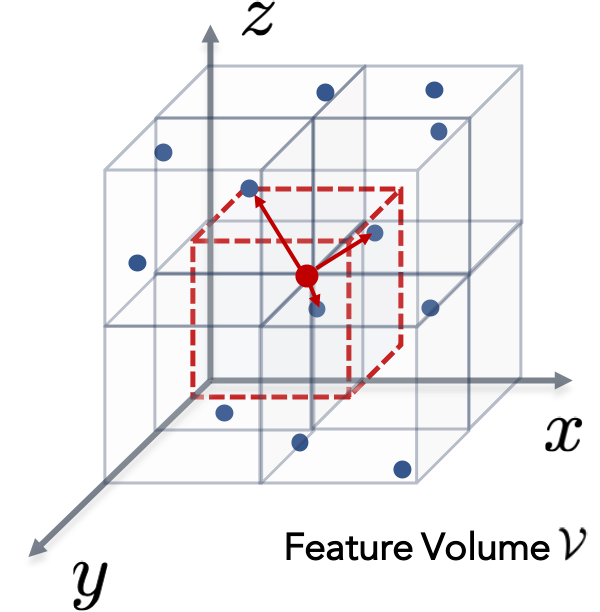}
  \caption{3D feature volume construction.}
\end{wrapfigure}
Given a point cloud $\mathcal{X}$, we first discretize the 3D space into a feature volume, $\mathcal{V}$, 
of resolution $H \times W \times D$. 
For each voxel center in $\mathcal{V}$, we then apply average pooling to aggregate features from surrounding points of $\mathcal{X}$. When there is no point near a voxel due to the sparsity of $\mathcal{X}$, that voxel remains empty. 
The point cloud $\mathcal{X}$ can be created from either single or multiple depth frames.

In our experiments, we build a hierarchical feature volume $\mathcal{V}$ with a resolution of [16, 32, 64]. Building a 3D hierarchical feature volume has been wildly used for recovering detailed 3D geometry, e.g.~\cite{chibane2020implicit, chen2021multiresolution}.
After processing the 3D feature volume with a 3D CNN, we use trilinear interpolation to get the feature of the query point $\textbf{p}$, \revise{which is sampled along the casting ray and} denoted as $\mathcal{V}(\mathbf{p})$. We use the drop-in replacement of \textit{grid\_sampler} from~\cite{wang2022go} to accelerate the training. 

\paragraph{Ray sampling strategy.}
Similar to~\cite{mildenhall2021nerf, wang2021neus}, we sample twice for each rendering ray.
First, we uniformly sample coarse points between the near bound $z_n$ and far bound $z_f$. 
Then, we use importance sampling with the coarse probability estimation to sample fine points. Folowing~\cite{wang2021neus}, the coarse probability is calculated based on $\Phi_h(s)$.
By this sampling strategy, our method can automatically determine sample locations and can collect more points near the surface, which makes the training process more efficient. 

\paragraph{Back projection}
Here we give details of the back projection function $\pi^{-1}$ to get point clouds from depth images.
Let $\textbf{K}$ be camera intrinsic parameters, $\xi=[\textbf{R}|\textbf{t}]$ be camera extrinsic parameters, where $\textbf{R}$ is the rotation matrix and $\textbf{t}$ is the translation matrix. 
$\bm{X}_{uv}$ is the projected point location and $\bm{X}_w$ is the point location in the 3D world coordinate. Then, according to the pinhole camera model:
\begin{equation}
    s \bm{X}_{uv} = \textbf{K} (\textbf{R} \bm{X}_w + \textbf{t}) \text{,}
\end{equation}
where $s$ is the depth value. After expanding the $\bm{X}_{uv}$ and $\bm{X}_{w}$:
\begin{equation}
    s \begin{bmatrix} u  \\ v \\ 1 \end{bmatrix} = \textbf{K} (\textbf{R} \begin{bmatrix} X  \\ Y \\ Z \end{bmatrix} + \textbf{t}) \text{.}
\end{equation}
Then, the 3D point location can be calculated as follows:
\begin{equation}
    \begin{bmatrix} X  \\ Y \\ Z \end{bmatrix} = 
    \textbf{R}^{-1} (\textbf{K}^{-1}s \begin{bmatrix} u  \\ v \\ 1 \end{bmatrix} - \textbf{t})
    \label{eq: back projection}
\end{equation}
The above Equation~\ref{eq: back projection} is the back-projection equation $\pi^{-1}$ used in this paper.

\paragraph{Training Time.} The \textbf{Ponder} model is \revise{pre-trained} with 8 NVIDIA A100 GPUs for 96 hours. 

\subsection{Transfer Learning Details}

\paragraph{3D scene reconstruction.}
ConvONet~\cite{peng2020convolutional} reconstructs scene geometry from the point cloud input. It follows a two-step manner, which first encodes the point cloud into a 3D feature volume or multiple feature planes, then decodes the occupancy probability for each query point. 
\revise{To evaluate the transfer learning capability of our point cloud encoder, we conduct an experiment where we replace the point cloud encoder of ConvONet directly with our pretrained encoder, without any additional modifications.}
We choose the highest performing configuration of ConvONet as the baseline setting, which uses a 3D feature volume with a resolution of 64.
For the training of ConvONet, we follow the same training setting as the released code\footnote{https://github.com/autonomousvision/convolutional\_occupancy\_networks}.

\paragraph{Image synthesis from point clouds.}
Point-NeRF~\cite{xu2022point} renders images from neural point cloud representation. It first generates neural point clouds from multi-view images, then uses point-based volume rendering to synthesize images. To transfer the learned network weight to the Point-NeRF pipeline, we 
1) replace the 2D image feature backbone with a pre-trained point cloud encoder to get the neural point cloud features, 
2) replace the color decoder by a pre-trained color decoder, 
3) keep the other Point-NeRF module untouched.
Since a large amount of point cloud is hard to be directly processed by the point cloud encoder, we down-sample the point cloud to 1\%, which will decrease the rendering quality but help reduce the GPU memory requirements. 
We report the PSNR results of the unmasked region as the evaluation metric, which is directly adopted from the original codebase\footnote{https://github.com/Xharlie/pointnerf}.
For training Point-NeRF, we follow the same setting as Point-NeRF.

\section{Supplementary Experiments}

\subsection{Transfer Learning}

\paragraph{Label Eﬃciency Training.}
\begin{figure}[t]
  \centering
   \includegraphics[width=0.8\linewidth, , trim=0 40 0 60]{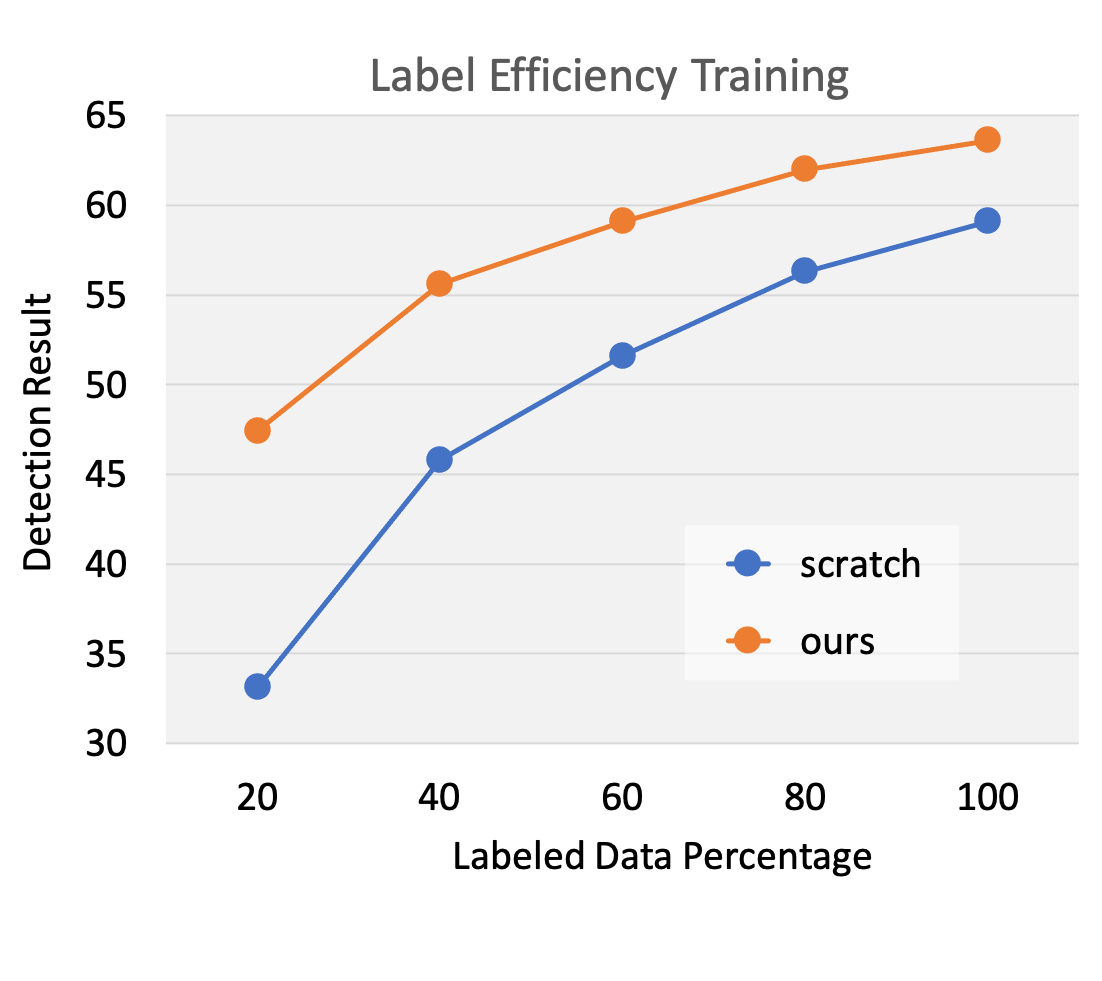}

   \caption{\textbf{Label efficiency training.} We show the 3d object detection experiment results using limited downstream data. Our pretrained model is capable of achieving better performance than training from scratch using the same percentage of data or requires fewer data to get the same detection accuracy. }
   \label{fig:label efficiency}
\end{figure}
We also do experiments to show the performance of our method with limited labeling for the downstream task. Specifically, we test the label efficiency training on the 3D object detection task for ScanNet. 
Following the same setting with IAE\cite{yan2022implicit}, we use 20\%, 40\%, 60\%, and 80\% of ground truth annotations. The results are shown in Figure \ref{fig:label efficiency}. 
We show constantly improved results over training from scratch, especially when only 20\% of the data is available. 

\paragraph{Color information for downstream tasks.}
Different from previous works, since our pre-training model uses a colored point cloud as the input, we also use color information for the downstream tasks. Results are shown in Table \ref{tab:supp detection}. 
Using color as an additional point feature can help the VoteNet baseline achieve better performance on the SUN RGB-D dataset, but get little improvement on the ScanNet dataset. This shows that directly concatenating point positions and colors as point features shows limited robustness to application scenarios.
By leveraging the proposed \textbf{Ponder} pre-training method, the network is well initialized to handle the point position and color features, and achieve better detection accuracy.

\begin{wraptable}{r}{0.23\textwidth}
    \centering
    \resizebox{0.2\textwidth}{!}{%
    \begin{tabular}{c| c c}
        \toprule
        Losses & $\rm{AP}_{50}$ $\uparrow$ & $\rm{AP}_{25}$ $\uparrow$ \\
        \toprule
        $L$ & \textbf{41.0} & 63.6 \\
        - $L_c$ & 40.9 & \textbf{64.2} \\
        - $L_d$ & 40.5 & 63.4 \\
        - $L_e$ & 40.9 & 63.3 \\
        - $L_e$ - $L_f$ & 40.7 & 63.1 \\
        - $L_e$ - $L_f$ - $L_s$ & 40.5  & 63.2 \\
        \bottomrule
    \end{tabular}}
    \vspace{5pt}
    \caption{\textbf{Ablation study for loss terms} $\textit{3D detection}$ $\textit{{AP}}_{25}$ and $\textit{{AP}}_{50}$ {on ScanNet.}}
    \vspace{-20pt}
    \label{tab:loss terms}
\end{wraptable}
\paragraph{Ablation study of different loss terms.}
The ablation study of different loss terms is shown in Tab.~\ref{tab:loss terms}, which demonstrates the effectiveness of each loss term.

 \begin{table*}
  \centering
  \begin{tabular}{c|c c c c |c c c c}
    \toprule
    \multirow{2}{*}{Method} & Detection & Pre-training & Pre-training & Pre-training & \multicolumn{2}{c}{ScanNet} & \multicolumn{2}{c}{SUN RGB-D}\\
     & Model & Type & Data & Epochs & $\rm {AP}_{50}$ $\uparrow$ & $\rm {AP}_{25}$ $\uparrow$ & $\rm {AP}_{50}$ $\uparrow$ & $\rm {AP}_{25}$ $\uparrow$\\
    \toprule
    VoteNet* & VoteNet* & - & - & - & 37.6 & 60.0 & 33.3 & 58.4 \\
    {DPCo}\cite{li2022invar3d} & VoteNet* & Contrast  & {Depth} & {120} & \textbf{41.5} & \textbf{64.2} & {35.6} & {59.8} \\
    {IPCo}\cite{li2022invar3d} & VoteNet* & Contrast & {Color \& Depth} & {120} & {40.9} & {63.9} & {35.5} & {60.2} \\
    \toprule
    VoteNet (w color) & VoteNet & - & - & - & 33.4 & 58.8 & 34.3 & 58.3 \\
    \textbf{Ponder} & VoteNet & Rendering & Depth  & 100 &  40.9 & \textbf{64.2} & 36.1 & 60.3 \\
    \textbf{Ponder} & VoteNet & Rendering & Color \& Depth  & 100 &  41.0 & 63.6 & \textbf{36.6} & \textbf{61.0} \\
    \bottomrule
  \end{tabular}
  \vspace{6pt}
  \caption{
  \textbf{3D object detection} $\textit{AP}_{25}$ \textit{and} $\textit{AP}_{50}$ {on ScanNet and SUN RGB-D.
  * means a different but stronger version of VoteNet.
  } 
  }
  \label{tab:supp detection}
  \vspace{-0.3cm}
\end{table*}

\paragraph{More comparisons on 3D detection.}
More detection accuracy comparisons are given in Table \ref{tab:supp detection}.
Even using an inferior backbone, our \textbf{Ponder} model is able to achieve similar detection accuracy with \ref{tab:supp detection} in ScanNet and better accuracy in SUN RGB-D.

\begin{table}[!htb]
    \begin{minipage}{.50\linewidth}
      \centering
        \resizebox{.85\linewidth}{!}{%
        \begin{tabular}{c| c c}
        \toprule
            Method & OA$\uparrow$ & mIoU$\uparrow$ \\
            \toprule
            DGCNN & 84.1 & 56.1 \\
            Jigsaw & 84.4 & 56.6 \\
            OcCo & 85.1 & 58.5 \\
            IAE & 85.9 & 60.7 \\
            \toprule
            \textbf{Ponder} & \textbf{86.2} & \textbf{61.1} \\
            \bottomrule
        \end{tabular}}
    \vspace{5pt}
    \caption{\textbf{3D semantic segmentation} \textit{OA and mIoU} on S3DIS dataset \texttt{with DGCNN model}. }
    \label{tab:3d semantic segmentation}
    \end{minipage}%
    \hfill
    \begin{minipage}{.43\linewidth}
      \centering
      \resizebox{.85\linewidth}{!}{%
        \begin{tabular}{c| c c}
            \toprule
            Epochs & $\rm{AP}_{50}$ $\uparrow$ & $\rm{AP}_{25}$ $\uparrow$ \\
            \toprule
            20 & 38.7 & 62.0 \\
            40 & 39.4 & 62.8 \\
            60 & 40.0 & 62.7 \\
            80 & 40.4 & 63.1 \\
            100 & \textbf{41.0} & \textbf{63.6} \\
        \bottomrule
        \end{tabular}}
    \vspace{5pt}
    \caption{
      \textbf{Ablation study for pre-training epochs}. 
  $\textit{3D detection}$ $\textit{{AP}}_{25}$ and $\textit{{AP}}_{50}$ {on ScanNet.}
      }
    \label{tab:pre-training epochs}
    \end{minipage} 
\end{table}

\paragraph{3D semantic segmentation with point-based approaches.}
Tab.~\ref{tab:3d semantic segmentation} shows our additional experiments with the point-based approach \textbf{Ponder}+DGCNN.

\paragraph{Ablation study of different pre-training epochs.}
Tab.~\ref{tab:pre-training epochs} shows that longer pre-training epochs lead to better performance in downstream tasks.

\subsection{More qualitative examples}
\begin{figure*}[htb]
  \centering
   \includegraphics[width=1.0\linewidth]{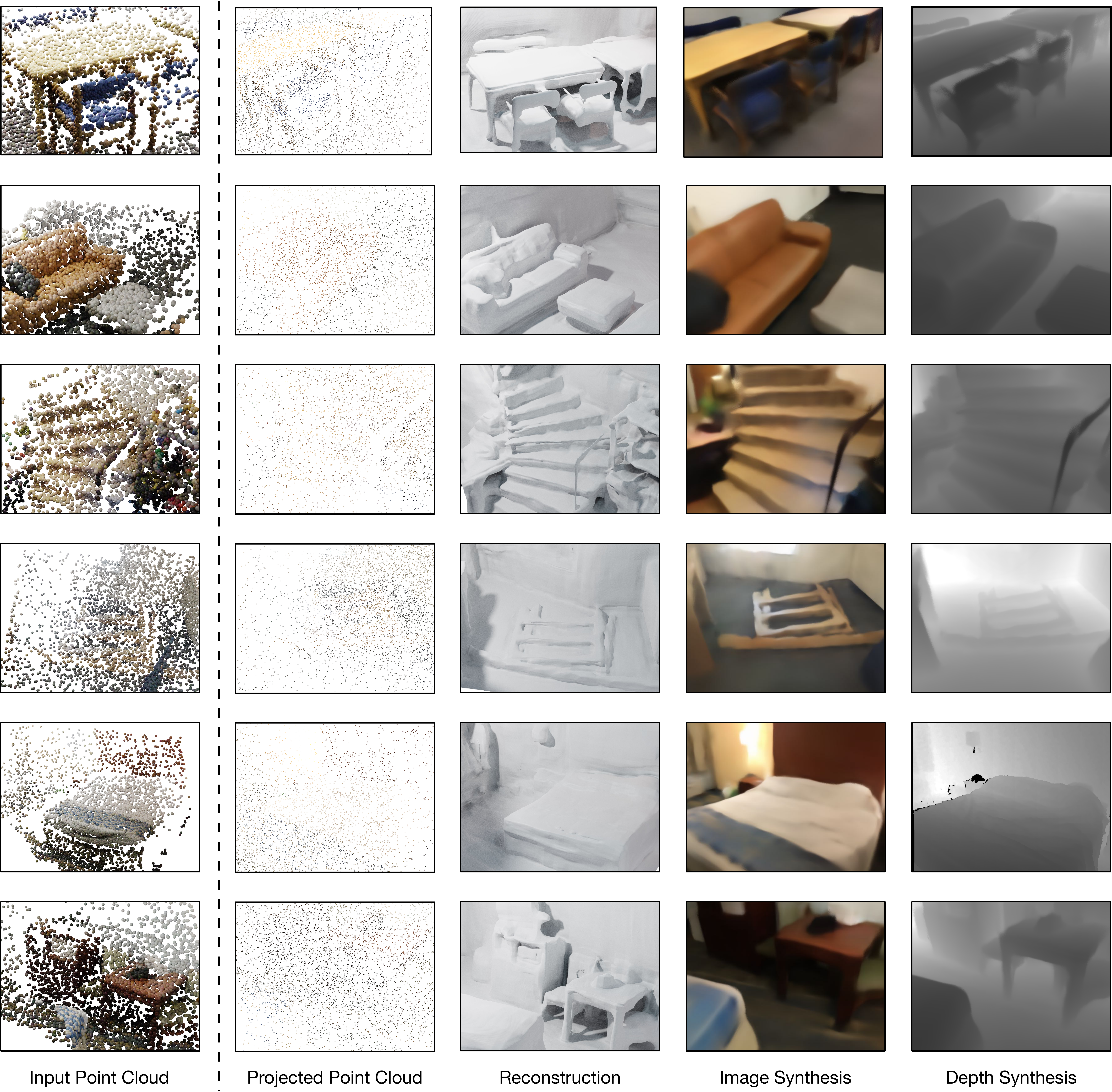}

    \caption{\textbf{More results of application examples of Ponder} on the ScanNet validation set (part 1). 
    The input point clouds are represented by large spheres for improved clarity. The projected point clouds illustrate the actual sparsity of the point data.
    }

   \label{fig:application 1}
   \vspace{-0.4cm}
\end{figure*}

\begin{figure*}[htb]
  \centering
   \includegraphics[width=1.0\linewidth]{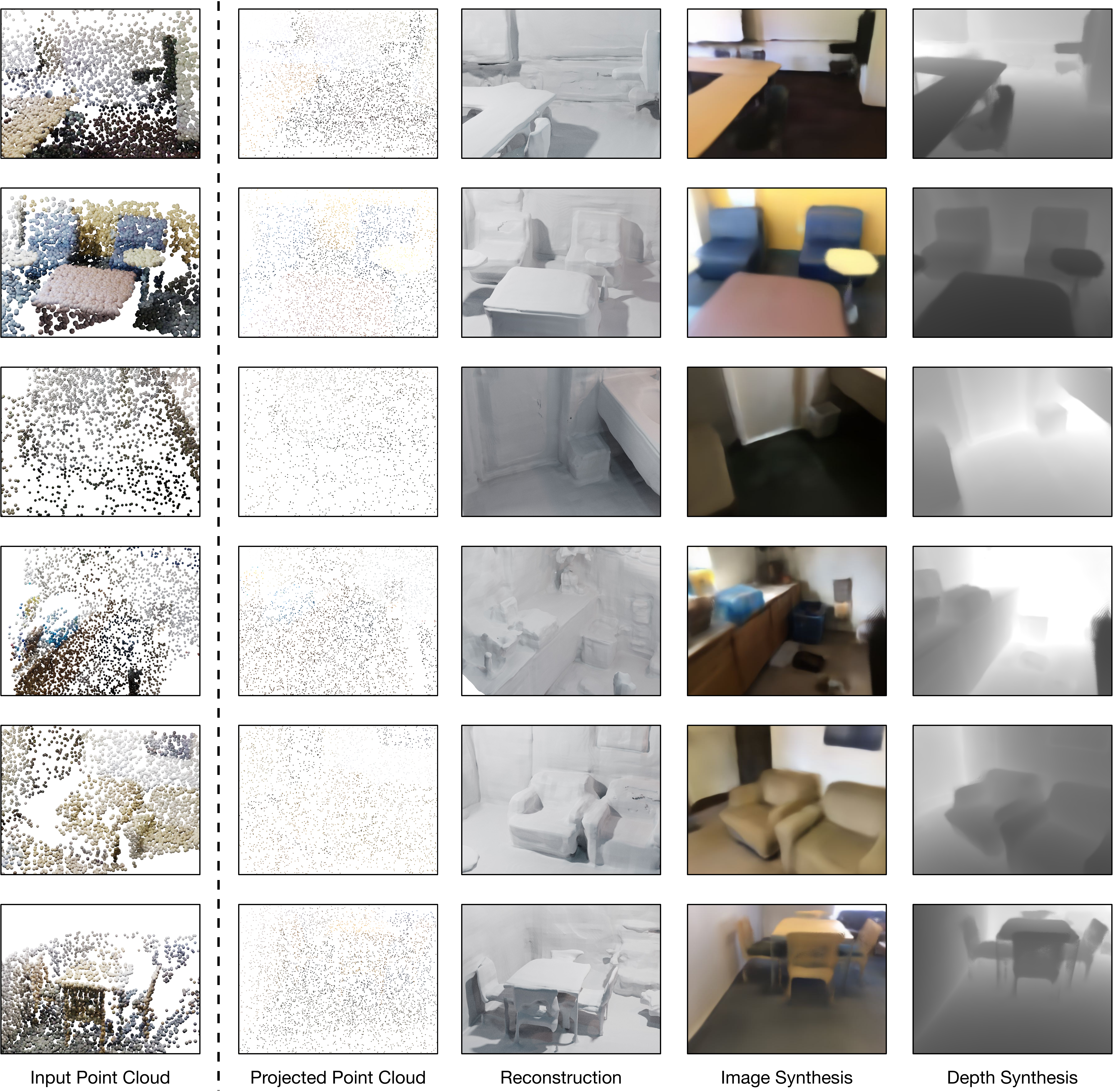}

   \caption{\textbf{More results of application examples of Ponder} on the ScanNet validation set (part 2). 
   The input point clouds are represented by large spheres for improved clarity. The projected point clouds illustrate the actual sparsity of the point data.
   }
   \label{fig:application 2}
   \vspace{-0.4cm}
\end{figure*}

As mentioned in the paper, the pre-trained \textbf{Ponder} model can be directly used for surface reconstruction and image synthesis tasks. 
We give more application examples in Figure \ref{fig:application 1} and Figure \ref{fig:application 2}.
The results show that even though the input is sparse point clouds from complex scenes, our method is able to recover high-fidelity meshes and recover realistic color and depth images.

\section{Multi-Camera 3D Object Detection}
To further verify the effect of utilizing rendering in self-supervised learning, we conduct exploratory experiments on the multi-camera 3D object detection task, which employs multiview images as input data.

\subsection{Experimental Setup}
\paragraph{Dataset.}
The nuScenes dataset~\cite{caesar2020nuscenes} is a popular benchmark for autonomous driving that includes data collected from six cameras, one LiDAR, and five radars. With 1000 scenarios, the dataset is split into three sets of 700, 150, and 150 scenes for training, validation, and testing, respectively.
The evaluation metrics used for 3D object detection in the nuScenes dataset incorporate the commonly used mean average precision (mAP) and a novel nuScenes detection score (NDS).

\paragraph{Implementation Details.}
For the downstream task, we adopt the latest state-of-the-art method, i.e., UVTR~\cite{li2022uvtr}, as our baseline.
Specifically, we use ResNet50-DCN~\cite{he2016deep, dai2017dcn} as the image backbone, which is initialized with the pre-trained weights (i.e., the weights of ResNet-50 Caffe model) from MMDetection\footnote{https://github.com/open-mmlab/mmdetection}.
To construct the 3D feature volume, we first project predefined 3D voxels to multi-view images through transformation matrices.
Then, the voxel features are interpolated from the image features via the projected pixel locations.
The resolution of the predefined 3D volume is $[128, 128, 5]$.
The model is trained with the AdamW optimizer with an initial learning rate of $2e^{-4}$ for $24$ epochs.

For pre-training, our model shares a similar architecture as the baseline, except that the point cloud is additionally used to supervise the rendered depth.
As our goal is to pre-train the 2D backbone, the point cloud is not used as input to construct the 3D feature volume, which is different from the process of Ponder in the main text.

\subsection{Main Results}

\begin{table}[htb]
  \centering
  \begin{tabular}{c c c c}
    \toprule
    Method & mAP$\uparrow$ & NDS$\uparrow$ \\
    \toprule
    UVTR\cite{li2022uvtr}  & 28.69 & 35.79 \\
    \textbf{Ours}   & \textbf{30.10} (\textcolor{purple}{+1.41}) & \textbf{36.31} (\textcolor{purple}{+0.52}) \\
    \bottomrule
  \end{tabular}
  \vspace{6pt}
  \caption{
  Performance comparisons on the nuScenes val set. 
  }
  \label{tab:supp outdoor_det}
\end{table}
\paragraph{The Effect of Pre-training.}
Table~\ref{tab:supp outdoor_det} shows that our method could yield up to 1.41\% mAP and 0.52\% NDS gains compared with the baseline, demonstrating the effectiveness of our pre-training method. The consistent improvement in both indoor and outdoor scenarios validates the robustness of our approach.

\begin{figure*}[htb]
  \centering
   \includegraphics[width=0.8\linewidth]{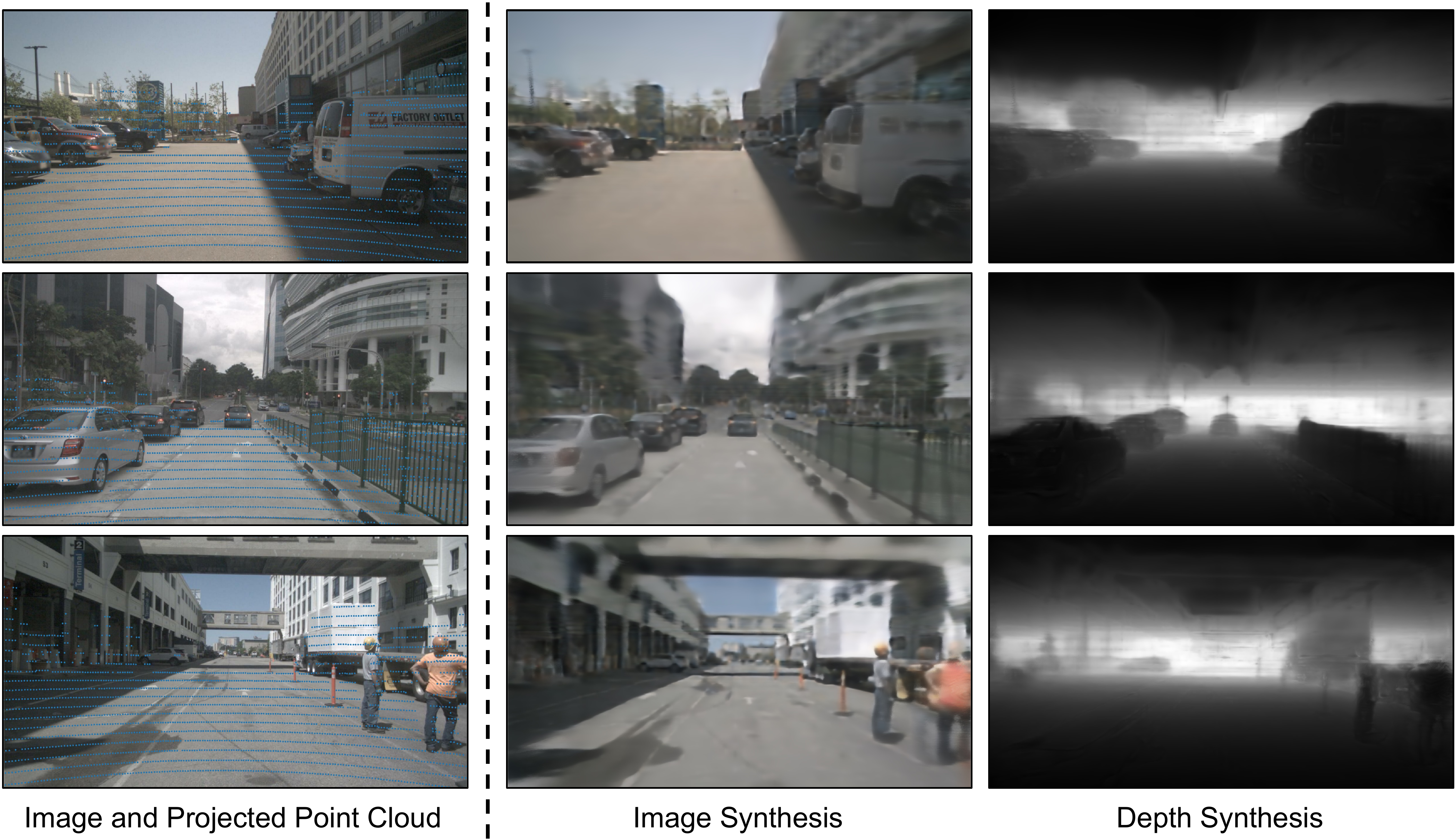}

   \caption{The predicted image and depth map on the nuScenes dataset. Left to right: image and projected point clouds, image predictions, and depth predictions. 
   }
   \label{fig:outdoor_render_vis}
   \vspace{-0.4cm}
\end{figure*}

\begin{figure*}[htb]
  \centering
   \includegraphics[width=0.9\linewidth]{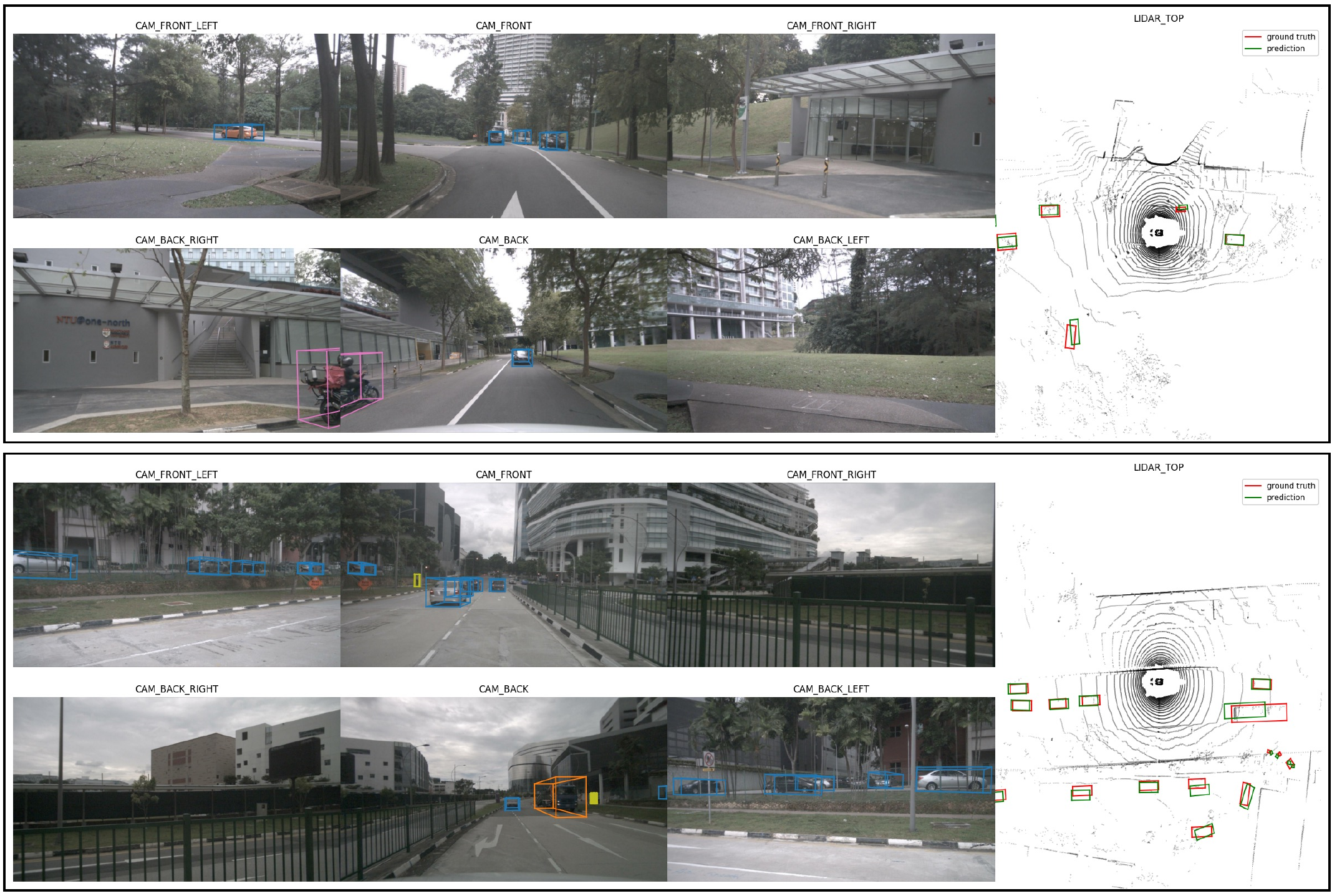}

   \caption{Qualitative results of multi-camera 3D object detection on the nuScenes dataset. We visualize the point cloud to better evaluate the quality of predicted bounding boxes.
   }
   \label{fig:outdoor_det_vis}
   \vspace{-0.4cm}
\end{figure*}

\paragraph{Visualization.}
Figure~\ref{fig:outdoor_render_vis} provides some qualitative results of the reconstructed image and depth map, which only takes the image as input during inference.
Our approach has the capability to estimate the depth of small objects, such as cars at a distance. This quality in the pre-training process encodes intricate and continuous geometric representations, which can benefit many downstream tasks.
In Figure~\ref{fig:outdoor_det_vis}, we present 3D detection results in camera space and BEV (Bird's Eye View) space.
Our model can predict accurate bounding boxes for nearby objects and also shows the capability of detecting objects from far distances.

\end{document}